\documentclass{article}

\usepackage{arxiv}

\usepackage[utf8]{inputenc} % allow utf-8 input
\usepackage[T1]{fontenc}    % use 8-bit T1 fonts
\usepackage{hyperref}       % hyperlinks
\usepackage{url}            % simple URL typesetting
\usepackage{booktabs}       % professional-quality tables
\usepackage{amsfonts}       % blackboard math symbols
\usepackage{nicefrac}       % compact symbols for 1/2, etc.
\usepackage{microtype}      % microtypography
\usepackage{lipsum}
\usepackage{fancyhdr}       % header
\usepackage{graphicx}       % graphics
\usepackage{multirow}
\usepackage[ruled,vlined]{algorithm2e}
\graphicspath{{media/}}     % organize your images and other figures under media/ folder
\usepackage{subcaption}
\usepackage[numbers]{natbib}
\title{WeChat-YATT: A Scalable, Simple, Efficient, and Production Ready Training Library
}

\author{
  Junyu Wu \\
  Tencent \\
  \texttt{nrwu@tencent.com} \\
   \And
  Weiming Chang \\
  Tencent \\
  \texttt{astrachang@tencent.com} \\
   \And
  Xiaotao Liu \\
  Tencent \\
  \texttt{xiaotaoliu@tencent.com} \\
   \And
  Guanyou He \\
  Tencent \\
  \texttt{guanyouhe@tencent.com} \\
  \And
  Tingfeng Xian \\
  Tencent \\
  \texttt{lucienxian@tencent.com} \\
  \And
  Haoqiang Hong \\
  Tencent \\
  \texttt{jeffhong@tencent.com} \\
   \And
  Boqi Chen \\
  Tencent \\
  \texttt{parkeychen@tencent.com} \\
    \And
  Hongtao Tian \\
  Tencent \\
  \texttt{oriontian@tencent.com} \\
    \And
  Tao Yang \\
  Tencent \\
  \texttt{luckytyang@tencent.com} \\
    \And
  Yunsheng Shi \\
  Tencent \\
  \texttt{yunshengshi@tencent.com} \\
    \And
  Feng Lin \\
  Tencent \\
  \texttt{flashlin@tencent.com} \\
    \And
  Ting Yao \\
  Tencent \\
  \texttt{tessieyao@tencent.com} \\
    \And
  Jiatao Xu \\
  Tencent \\
  \texttt{sunnyxu@tencent.com} \\
}

\begin{document}
\maketitle

\begin{abstract}

  Reinforcement Learning from Human Feedback (RLHF) has emerged as a prominent paradigm for training large language models and multimodal systems. Despite the notable advances enabled by existing RLHF training frameworks, significant challenges remain to scale to complex multimodal workflows and adapt to dynamic workloads. In particular, current systems often encounter limitations related to controller scalability when managing large models, as well as inefficiencies in orchestrating intricate RLHF pipelines, especially in scenarios that require dynamic sampling and resource allocation. In this paper, we introduce \textbf{WeChat-YATT} (\textbf{\textit{Y}}et \textbf{\textit{A}}nother \textbf{\textit{T}}ransformer \textbf{\textit{T}}rainer in \textbf{\textit{WeChat}}), a simple, scalable, and balanced RLHF training framework specifically designed to address these challenges. WeChat-YATT features a parallel controller programming model that enables flexible and efficient orchestration of complex RLHF workflows, effectively mitigating bottlenecks associated with centralized controller architectures and facilitating scalability in large-scale data scenarios. In addition, we propose a dynamic placement schema that adaptively partitions computational resources and schedules workloads, thereby significantly reducing hardware idle time and improving GPU utilization under variable training conditions. We evaluate WeChat-YATT across diverse experimental scenarios, demonstrating its substantial throughput improvements over state-of-the-art RLHF training frameworks. Furthermore, WeChat-YATT has been successfully deployed to train models that support WeChat product features for a large-scale user base, underscoring its effectiveness and robustness in real-world applications. We have made WeChat-YATT publicly available at \url{https://www.github.com/tencent/WeChat-YATT}.

\end{abstract}


\section{Introduction}

Transformer-based large models and their multimodal extensions have achieved remarkable progress, demonstrating state-of-the-art performance across a wide range of natural language processing, understanding, and generation tasks \cite{gpt4, llama3, dseekv3, googlegemini, flux, gpt3survey, sd, kimi, attn}. These models, characterized by their massive parameter counts and ability to process extremely long context sequences, have become the foundation for many advanced AI applications. However, the unprecedented scale of LLMs has introduced significant challenges for efficient training, and traditional data-parallel training strategies are often insufficient to handle the memory and computational demands of such large models. To address these challenges, the research community has developed a variety of hybrid parallelism techniques \cite{mlm1, mlm2, mlm3, mlm4, ringattn}. These approaches enable the training of models that would otherwise exceed the memory and compute capabilities of individual devices, by partitioning both the model and the data across multiple GPUs or nodes.

The integration of Reinforcement Learning from Human Feedback (RLHF) has significantly transformed the training paradigms of large language models \cite{qwen25math, dseekr1, dspo}. By embedding human preferences into the optimization process, RLHF requires the coordination of multiple models—such as reward models, policy models, and reference models—within a unified training pipeline \cite{ppo, grpo, remax}. This multi-model architecture substantially amplifies the number of parameters and the volume of intermediate activations that must be handled, thereby straining the capabilities of current distributed training infrastructures. The scalability challenges are further exacerbated in multimodal RLHF scenarios. The incorporation of diverse data modalities (e.g., text, images, audio) not only increases the heterogeneity and volume of training data but also intensifies the demands on inter-node communication and memory bandwidth. As a result, the system faces pronounced communication overheads and memory bottlenecks, hindering its ability to scale efficiently to larger datasets and model sizes. Addressing these bottlenecks is critical for facilitating the practical deployment of RLHF in increasingly complex and data-rich environments.

Sample efficiency is a critical factor in the RLHF training process, as it directly influences convergence rates and overall model performance \cite{ds_byte}. While flexible sampling strategies have been shown to accelerate convergence, most existing RLHF training frameworks lack robust support for efficient and adaptive sampling mechanisms. Although hybrid parallelism theoretically enables substantial scalability, practical implementations often encounter significant communication overhead as the degree of parallelism increases. Specifically, as the number of models and the length of context sequences expand, the costs associated with synchronizing parameters and activations across distributed devices can escalate rapidly, thereby diminishing the anticipated benefits of parallelization. To address these challenges, recent work such as \citet{nemoaligner} has conceptualized the RLHF training workflow as a control flow problem, employing techniques such as multi-cluster, multi-model partial colocated, and memory swapping to alleviate peak GPU memory consumption. Similarly, other systems \cite{openrlhf,verl} have explored alternative strategies, with \citet{verl} further enhancing throughput by integrating both single- and multi-controller architectures alongside statically optimized placement schemes. These advancements underscore the ongoing need for innovative approaches to improve sampling efficiency and resource utilization in large-scale RLHF training.

\begin{enumerate}
  \item \textbf{Scalability bottlenecks in large-scale multimodal scenarios:} The management of extensive multimodal data (e.g., images, videos) can result in a single controller becoming a communication or memory bottleneck, thereby constraining system throughput or even causing operational failures.
  \item \textbf{Inefficiency under dynamic sampling and generative reward modeling:} Workflows that require frequent dynamic sampling or generative reward computation are susceptible to significant overhead from model swapping and long-tail effects (i.e., straggler tasks), which can substantially reduce hardware utilization and training efficiency.
  \end{enumerate}
  
  To address these challenges, we introduce \textbf{WeChat-YATT}, a novel RLHF training framework designed to enhance scalability, flexibility, and efficiency in large-scale, multi-model training environments. The key contributions of this work are as follows:
  
  \begin{enumerate}
  \item \textbf{Parallel Controller Programming Model:} We propose a parallel controller programming model that facilitates the construction of complex RLHF workflows. This model supports efficient distributed execution and flexible control flow, effectively mitigating the bottlenecks associated with centralized controller architectures.
  \item \textbf{Dynamic Scaling Placement Schema:} We present a dynamic scaling placement schema that enables efficient generative rewarding and dynamic sampling. By adaptively partitioning resources and supporting fine-grained scheduling, our approach minimizes device idle time (“bubbles”) and optimizes hardware utilization, even under highly variable workloads.
  \end{enumerate}

The proposed WeChat-YATT advances the state-of-the-art in RLHF training by effectively addressing the critical bottlenecks of controller scalability and resource placement. We hope that WeChat-YATT will serve as a robust foundation for both academic research and industrial applications, fostering further innovation and collaboration within the community.

\section{Background}

\subsection{Distributed Training of Deep Models}

Data Parallelism involves splitting the training data across multiple devices (such as GPUs), where each device holds a copy of the model and processes a different mini-batch of data. Gradients are averaged (or summed) across devices to update the model parameters synchronously \cite{torchddp, horovod}. Tensor parallelism splits individual layers or tensors of the model across multiple devices. Each device is responsible for computing a portion of the operations within a layer, enabling the training of very large models that cannot fit into a single device’s memory \cite{mlm1}. Pipeline parallelism divides the model into sequential stages, with each stage placed on a different device. Mini-batches are divided into micro-batches and passed through the pipeline, allowing different devices to work on different parts of the input simultaneously, improving hardware utilization \cite{mlm2, gpipe, pipedream}.

ZeRO is an optimization technique that partitions model states (such as optimizer states, gradients, and parameters) across devices to minimize memory redundancy. This enables the training of much larger models by reducing the memory consumption per device \cite{zero3, zero1}.

Context Parallelism is a parallelization strategy where input context is distributed across multiple devices. This allows for efficient processing of long sequences or large batches by splitting the workload. Techniques like ring attention are examples of context parallelism, enabling scalable and memory-efficient attention computation across devices \cite{llama3, ringattn, ulysess, fang2024uspunifiedsequenceparallelism}.

Expert Parallelism is used in Mixture-of-Experts (MoE) models, where different subsets of the model (experts) are distributed across devices. During training and inference, only a subset of experts is activated for each input, allowing efficient scaling and utilization of resources \cite{mlm4, tutel}.

\subsection{RLHF Workflow}

RLHF aligns the linguistic space of LLMs with human preferences, using a set of human-ranked candidates of given prompts \cite{ppo, bai2022traininghelpfulharmlessassistant}. An typical reinforcement learning trainer consists of multiple models, e.g., an actor, a critic, a reference policy, and one or multiple reward models. The critic and reward models are usually models tuned in the human preference dataset, with the language modeling head replaced by a numerical output head (Bradley-Terry reward model) \cite{bai2022traininghelpfulharmlessassistant, ouyang2022traininglanguagemodelsfollow, btrm, rethinkbt}. Bradley-Terry reward models can be replaced with generative models, which obtain reward scores through token prediction; furthermore, leveraging chain-of-thought reasoning, the accuracy and final performance of the actor model can be further improved \cite{genrm, cot}.

\begin{figure}
    \begin{subfigure}{.45\textwidth}
        \centering
        \includegraphics[width=.8\linewidth]{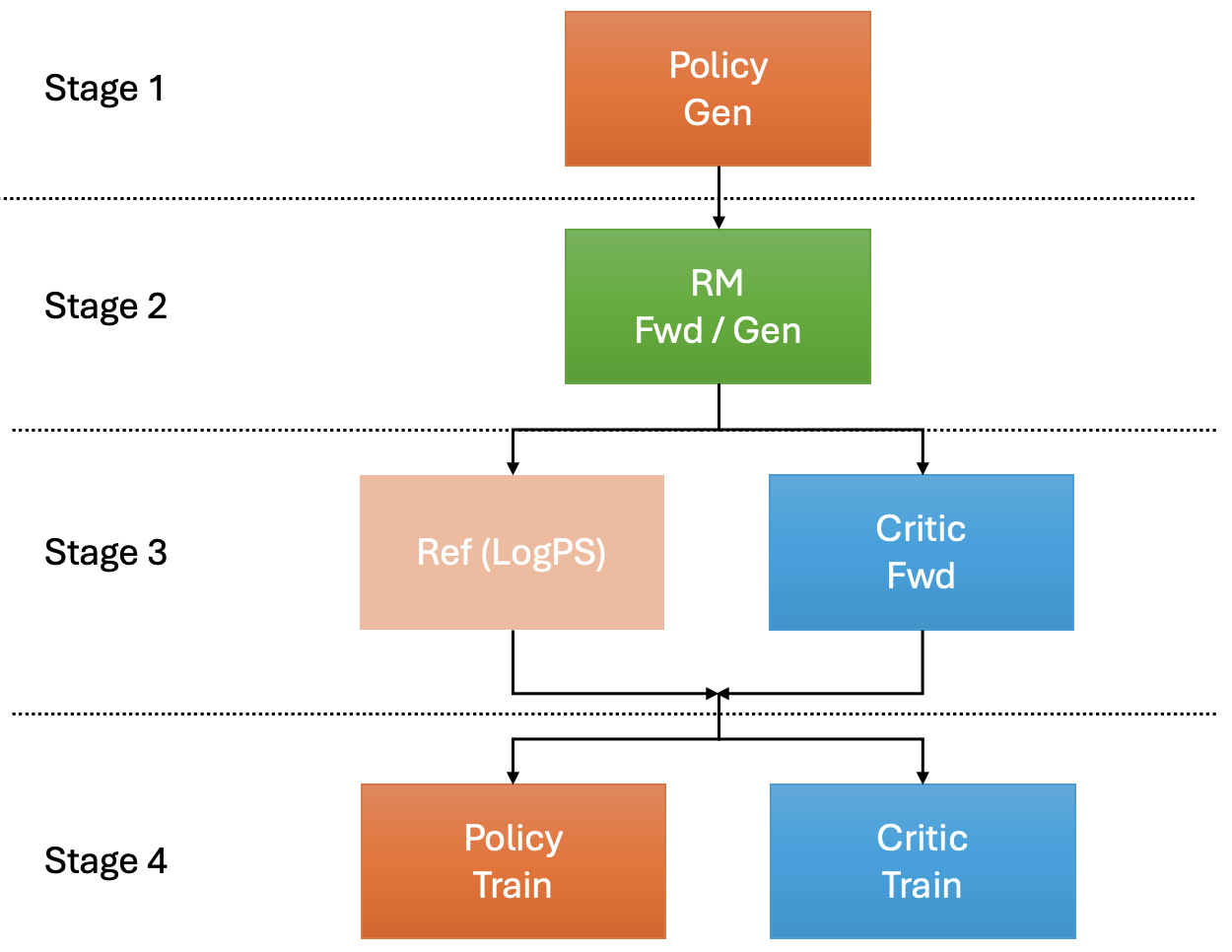}
        \caption{Regular RLHF workflow}
        \label{fig:workflow1}
    \end{subfigure}
    \begin{subfigure}{.45\textwidth}
        \centering
        \includegraphics[width=.8\linewidth]{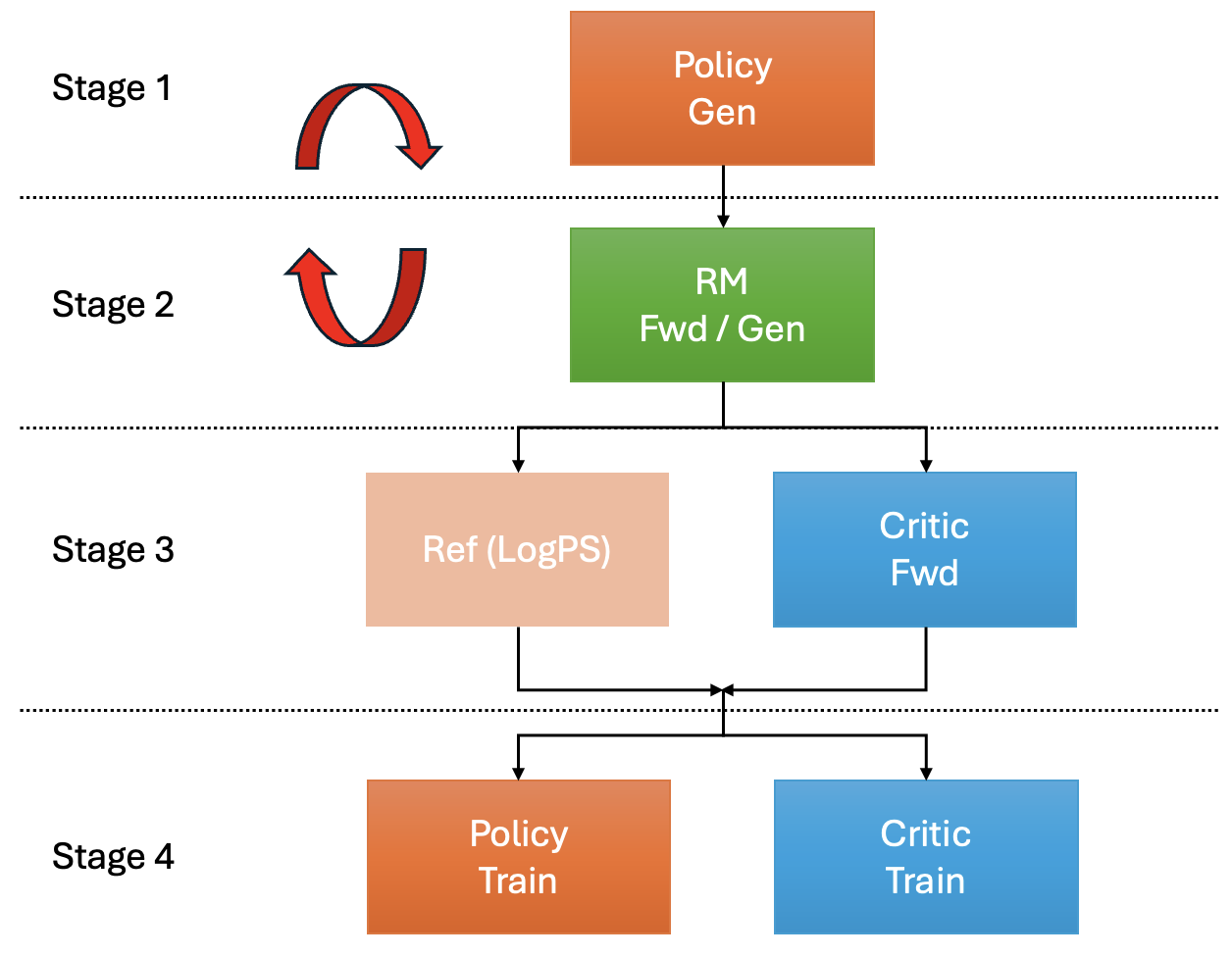}
        \caption{RLHF workflow with Dynamic Sampling}
        \label{fig:workflow2}
    \end{subfigure}
    \caption{RLHF workflow}
    \label{fig:workflow}
\end{figure}

Most RLHF workflows largely follow this 4-stage workflow \cite{ppo, grpo, dapo, dseekr1, qwen25math, qwen3, remax}(Sometimes, the first three stages are referred to as 'rollout' (Figure \ref{fig:workflow1}):

\begin{enumerate}
  \item Stage 1(\textbf{Generation}): The actor produces responses from a batch of prompts using auto-regressive generation.
  \item Stage 2(\textbf{Rewarding}): The reward model computes rewards for generated responses.
  \item Stage 3(\textbf{Preparation}): Using generated responses and rewards, the critic computes their values, the policy and reference policy compute their reference log probabilities.
  \item Stage 4(\textbf{Training}): The actor and the critic are trained on the batch of data produced by previous stages and the loss function.
\end{enumerate}

In engineering practice, the generation step is typically implemented using a specialized language model inference engine (e.g., vLLM \cite{vllm}, sglang \cite{sglang}) \cite{openrlhf, nemoaligner, verl}.

To simplify implementation, most systems use a single controller to manage data transfer, intermediate results, and workflow orchestration through a central point \cite{openrlhf, verl}. However, this centralized approach can sometimes become a bottleneck, especially when the trainer itself has high memory requirements.

\subsection{RLHF Placement}

Workflow is at the core of RLHF training, and model placement is a crucial component of the RLHF workflow. Currently, there are at least two main approaches:
\begin{itemize}
    \item \textbf{Co-exist:} Different models are assigned to separate groups of devices, allowing them to run in parallel as long as there are no data dependencies between them.
    \item \textbf{Co-locate:} Multiple models are placed on the same set of GPUs, sharing GPU memory. To avoid out-of-memory (OOM) errors, which can easily occur if large models are run simultaneously, these colocated models are executed one after another in a time-sharing fashion.
\end{itemize}

Although \citet{verl} proposed methods for calculating placement based on workload, both the open source implementations of OpenRLHF \cite{openrlhf} and VeRL \cite{verl} adopt a co-locate all models strategy. The partial colocated approach introduces overhead from swapping among multiple models. For example, swapping a 32B model could take nearly a minute, and updating weights may take tens of seconds. In typical GRPO training, this overhead is not a major issue, since the primary bottlenecks in RLHF training are the generation and the forward/backward passes of modeling. Compared to the tens of minutes spent on rollout and training, model swapping is not the system bottleneck.

Some works \cite{areal, flexrlhf} have adopted a coexisting approach, which effectively avoids the overhead of model swapping. However, to mask device idle time during workflow state transitions, these methods introduce varying degrees of asynchronous overlap between the rollout and training stages. This can lead to suboptimal model convergence; for example, \citet{areal} introduces staleness-aware training to address this issue. In our experience, many algorithm engineers find that this trade-off adds extra cognitive burdens.

\section{Design of WeChat-YATT}
\subsection{Overview}

\begin{figure}
      \centering
      \includegraphics[width=.85\linewidth]{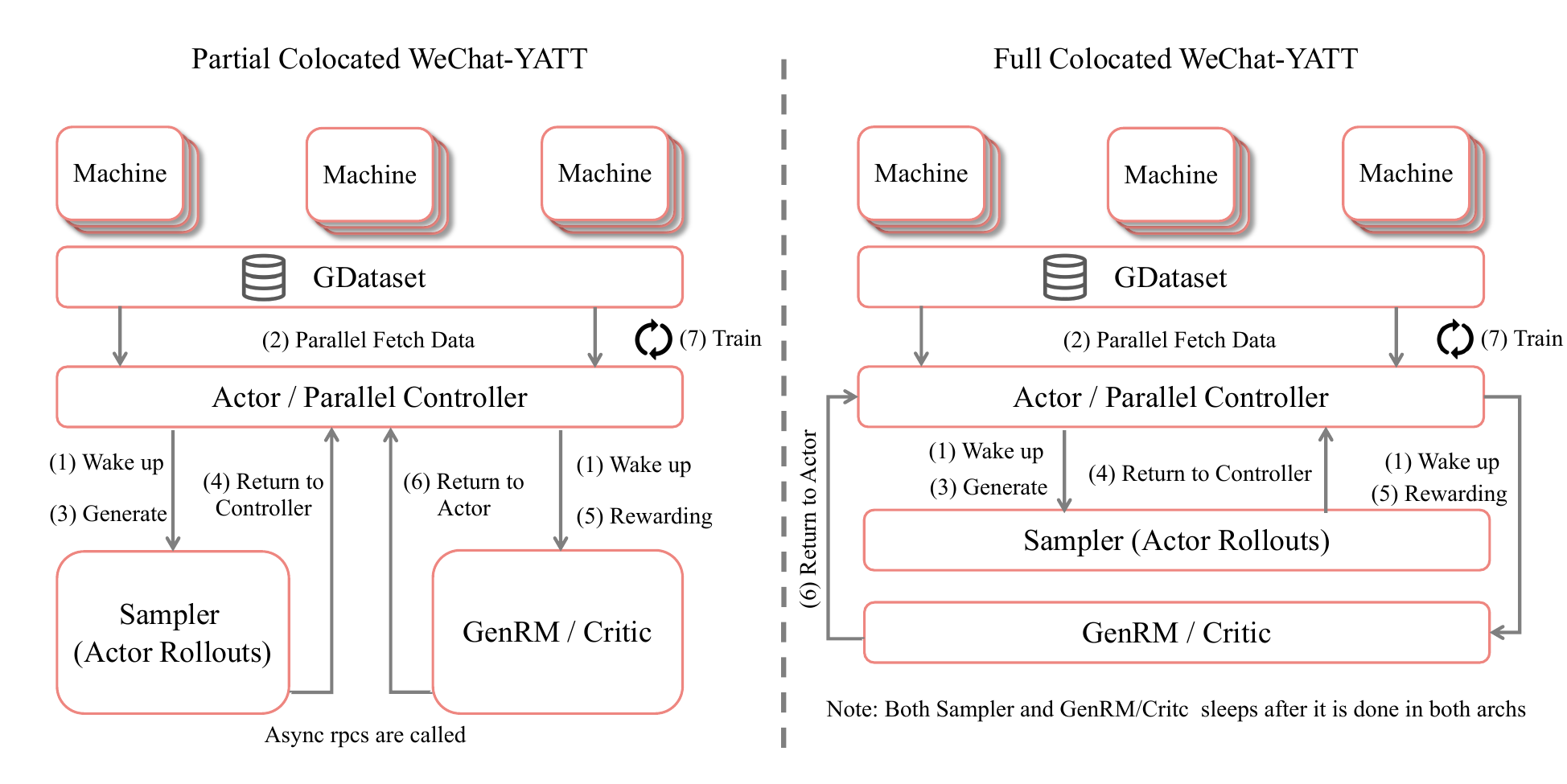}
      \caption{Overview of partial colocated WeChat-YATT and colocated WeCaht-YATT}
    \label{fig:overview}
\end{figure}

Figure \ref{fig:overview} presents a high-level overview of the proposed WeChat-YATT framework. WeChat-YATT supports two distinct resource placement architectures: partial colocated and full colocated.

In the partial colocated architecture, the \textit{Sampler} and the \textit{GenRM} (Generative Reward Model) components are deployed independently and interact asynchronously. The allocation of computational resources to these two components directly influences throughput during the rollout phase, a placement method that will be examined in detail in Section \ref{sec:dp}. Within this architecture, the actor model leverages all available machine resources across the cluster, while parallel controllers correspond to the ranks of model-parallel worker heads associated with the \textit{Actor}. The training loop is initiated by wake-up remote procedure calls (RPCs), activating both the \textit{Sampler} and \textit{GenRM}. Subsequently, parallel controllers retrieve data from the distributed \textit{GDataset}, thereby mitigating excessive memory consumption on any single machine. Notably, both the \textit{Sampler} and \textit{GenRM} operate asynchronously: upon completing each microbatch, the \textit{Sampler} immediately transmits the generated data to the \textit{GenRM} without impeding ongoing generation processes. Once the rollout phase concludes, the \textit{Actor} instructs other components to enter a sleep state, offloads their memory, and subsequently loads the actor model to initiate the training phase.

In contrast, the full colocated architecture adopts a more integrated approach, wherein all components share the resources of all machines. Under this configuration, the \textit{GenRM} must await the \textit{Sampler}'s completion of the entire sample generation process. Following this, the \textit{Sampler} offloads its parameters to disk, and the \textit{GenRM} is loaded into memory to initiate the reward computation process. Upon completing reward computation, the \textit{Actor} is loaded into memory, while the \textit{GenRM} is offloaded. This workflow is inherently synchronous, with transitions between components occurring only after the preceding process is fully completed. Consequently, in scenarios characterized by frequent long-tail effects, the full colocated architecture may not offer optimal performance.

Overall, WeChat-YATT’s flexible resource placement strategies enable efficient orchestration of RLHF workflows, accommodating diverse operational requirements and workload characteristics.

\subsection{Parallel Controllers}

While hybrid controllers offer notable advantages in terms of programming convenience, they may lack the necessary flexibility or introduce system bottlenecks under specific conditions. Two principal challenges inherent to both single and hybrid controller paradigms merit attention.

\textbf{Resource Bottlenecks in Data-Intensive Workflows.} A significant limitation arises when large-scale features—such as extensive collections of images or videos, particularly in image generation tasks \cite{flux, sd, dancegrpo}—must be transferred within the control flow. In such cases, the memory capacity and RPC network bandwidth of a single controller can quickly become limiting factors. Additionally, the execution of complex procedures within the controller may result in CPU bottlenecks. For instance, in specialized algorithmic scenarios, even with a controller provisioned with 2048 GB of memory, practical constraints such as model and optimizer swap buffers typically reduce available memory by half. Conducting a rollout of 1024 samples, each comprising 32 images at 2K resolution, would already consume approximately 768 GB of memory. Further operations, including data copying, increased video frame counts, or communication buffer overhead, can precipitate out-of-memory (OOM) errors. Employing shallow copies can mitigate memory consumption by preventing redundant tensor creation during repeated sampling. Nevertheless, if the sampling algorithm incorporates advanced data augmentation strategies, memory usage may escalate once again.

\textbf{Limited Granularity in State Transitions.} From a systems architecture perspective, the hybrid controller essentially functions as a single entity managing proxy objects for GPU resource pools associated with multiple roles (e.g., policy, reward, reference, critic). This design restricts the controller to stage-wise transitions, thereby impeding the realization of local state changes. The inability to perform localized state transitions complicates support for specialized sampling processes, such as dynamic sampling or reward-augmented generation \cite{dapo, rad}.

\begin{figure}
    \begin{subfigure}{.5\textwidth}
      \centering
      \includegraphics[width=.6\linewidth]{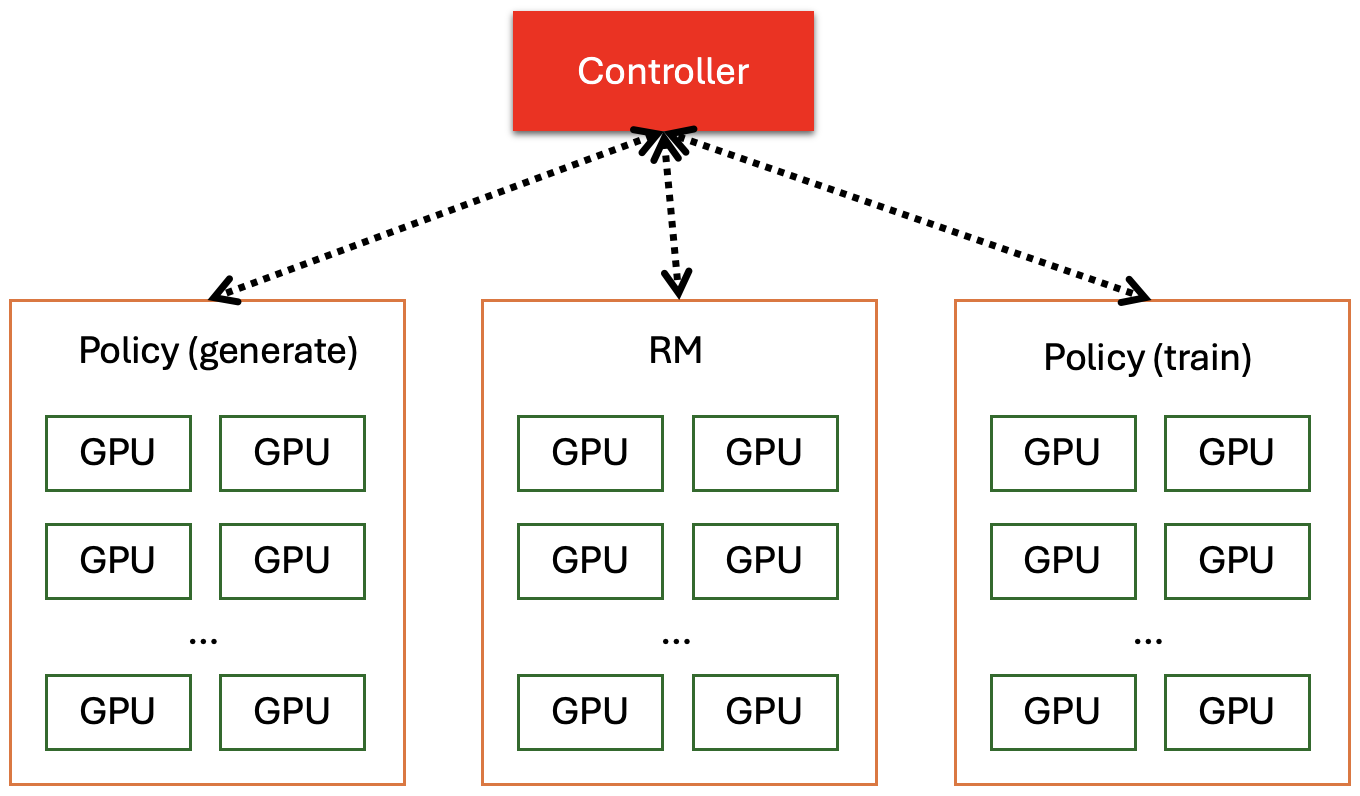}
      \caption{Hybrid Controller}
      \label{fig:sfig1}
    \end{subfigure}%
    \begin{subfigure}{.5\textwidth}
      \centering
      \includegraphics[width=.9\linewidth]{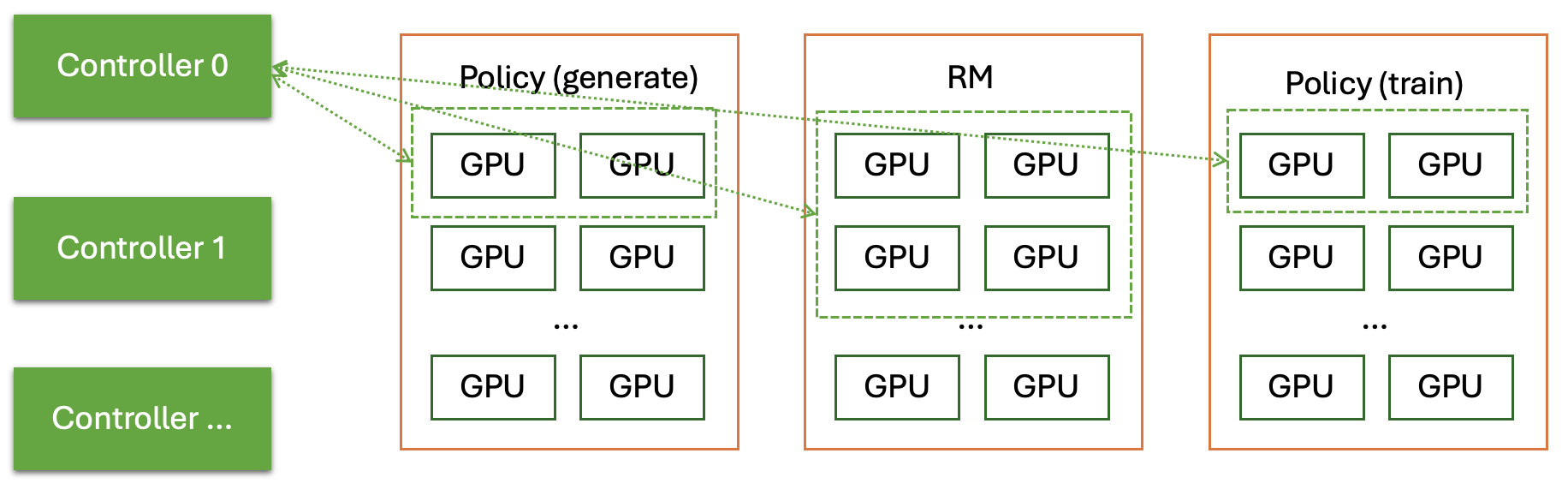}
      \caption{Parallel Controllers}
      \label{fig:sfig2}
    \end{subfigure}
\caption{Each controller in parallel controllers is only responsible for part of workload and resources, avoids overwhelming of controller bottleneck resources}
\label{fig:3l-controller}
\end{figure}

To address these challenges, we introduce a parallel controller architecture. As illustrated in Figure \ref{fig:3l-controller}, reinforcement learning (RL) tasks are partitioned using the conventional SPMD (Single Program Multiple Data) paradigm, wherein multiple controllers independently manage distinct data subsets. Statistically, as batch size increases, the distribution of workload among controllers becomes increasingly balanced. This effect is underpinned by the law of large numbers, which ensures that, with sufficiently large sample sizes, the data allocated to each controller converges toward its expected value, thereby promoting equitable load distribution.

In the parallel controller architecture, each controller is responsible for managing a subset of system resources, with individual resources potentially managed by either a single controller or shared among multiple controllers. This design enables the concurrent existence of multiple operational stages within the system, thereby facilitating fine-grained state transitions that can be tailored to the requirements of specific reinforcement learning algorithms.

Coordinating multiple reinforcement learning roles using several controllers is often considered a complex undertaking. In our proposed framework, however, we allocate distinct controllers to manage different RL roles. This approach mirrors the hybrid controller paradigm \cite{verl}, wherein each actor internally employs a multi-controller structure to enhance computational efficiency \cite{evalgpuinter, vllm}.

The parallel controller architecture can be interpreted as a pragmatic evolution of the hybrid controller, designed to provide increased controller resources and enable more flexible resource scheduling.  For programming efficiency, our experience suggests that, the adoption of a multi-controller architecture does not pose substantial programming difficulties. In our implementation, we further decompose the top-level controller and employ collective communication mechanisms to facilitate coordination among controllers. Within each worker cluster managed by a controller, we preserve a control pattern analogous to that of the hybrid controller, thereby maintaining operational consistency while improving scalability and flexibility.

\subsection{Dynamic Placement \label{sec:dp}}
% \subsubsection{Generative Rewarding and partially colocated}

The generative rewarding approach has demonstrated considerable effectiveness in recent studies \cite{genrm, genverifier}. In contrast to discriminative LLM-based verifiers, which do not leverage the generative capabilities of pretrained LLMs, generative rewarding utilizes standard next-token prediction to assess solution correctness. Specifically, this method represents correctness through the LLM’s probability distribution over tokens, rather than by predicting a separate numerical score. This design preserves the generative abilities of the language model, as the verification decision is formulated as an additional token prediction. Furthermore, it enables several inherent advantages of LLMs, including unified training for both solution generation and verification, support for chain-of-thought reasoning, and the ability to perform inference-time computation \cite{genverifier}.

To implement this, we adopt a partial colocation strategy, which, as previously discussed, offers an efficient solution without introducing additional algorithmic trade-offs. In our framework, a causal text generation inference engine replaces the traditional regression-based rewarding model. Upon completion of the generation phase, the policy model is offloaded to CPU memory, and the generative reward model is loaded onto GPU memory. This reward model then generates reward scores through a combination of text generation and regular expression matching.

Empirically, the partial colocation approach has proven highly effective, resulting in minimal hardware idle time. When compared to the relatively processes of text generation, reward computation, and model training, the overhead associated with swapping models in and out of GPU memory is negligible. For example, swapping a 32B model typically requires only 20 seconds for SGLang, which constitutes a minor fraction of the total computation time.

However, dynamic sampling scenarios introduce additional complexity. As described in \citet{dapo}, prompts with accuracy scores of 1 or 0 (i.e., batches with uniformly high or low rewards) are filtered out, triggering resampling. This process necessitates additional rounds of generation and reward computation, as illustrated in Figure \ref{fig:workflow2}.

\begin{enumerate}
\item \textbf{Frequent Swapping Overhead.} As training progresses and model performance improves, the acceptance rate of generated samples tends to decrease. This results in more frequent resampling and, consequently, more frequent model swaps between the policy and reward models. Under these conditions, the previously negligible model swapping overhead can accumulate, potentially becoming a bottleneck in the training pipeline and limiting overall system efficiency.
\item \textbf{Long-tail Accmulation.} During the generation phase, the occurrence of long-tail outputs is inevitable. These outputs can reduce overall GPU cluster utilization. In partial colocation settings, where frequent transitions between multiple stages are required, the long-tail effect may be further exacerbated. Specifically, the uneven completion times of different tasks can lead to increased resource fragmentation and idle GPU time, thereby posing additional challenges to maintaining high cluster efficiency.
\end{enumerate}

To address the challenges associated with model swapping overhead and the exacerbated long-tail effect, we develope a dynamic placement strategy that integrates dynamical resource allocation in the training process under patial colocated WeChat-YATT paradigm.

In stages 1 and 2, the simultaneous residency of both the policy generation model and the rewarding model in GPU memory eliminates the need for model swapping, thereby avoiding the memory copy and graph capturing overhead typically incurred during dynamic sampling. Furthermore, by maintaining the policy model in memory following the reward computation, subsequent rounds of generation can be initiated immediately. This finer-grained control effectively reduces idle periods ("bubbles") during the long-tail phase of the system, thereby enhancing overall hardware utilization and system efficiency.

From an engineering standpoint, dynamically partitioning the GPU cluster according to workload is technically feasible. However, this approach does not fully account for the fact that system load can fluctuate due to algorithmic factors \cite{dseekr1}. During reinforcement learning, for example, the average response length of models on the training set tends to increase as the model learns to solve more complex reasoning tasks, often requiring extended "thinking time." This trend is consistently observed throughout the training process, as models progressively acquire the capability to address increasingly sophisticated reasoning challenges by leveraging longer test-time computations. Such computations may involve generating responses that range from hundreds to thousands of reasoning tokens, thereby enabling the model to explore and refine its reasoning processes in greater depth. Our simulation experiments have frequently exhibited phenomena consistent with these observations.

When the workload is dynamic, static estimation methods such as those proposed in \citet{verl} become inadequate for determining optimal resource placement. It is inherently difficult to predict how the output length of the model will evolve over the course of training. Moreover, as the model's reasoning responses become longer, it becomes increasingly challenging to anticipate the corresponding generation length required for reward computation. This unpredictability further complicates the design of efficient placement strategies in dynamic training environments.

To mitigate the aforementioned challenges, we propose a dynamic placement strategy. At the onset of training, initial resource allocation between the policy model and the generative rewarding model is determined using simple heuristic methods. As training progresses, hardware utilization is continuously monitored, enabling the gradual reduction of resources assigned to roles exhibiting low utilization and the reallocation of these resources to roles with higher demand. This adaptive approach ultimately leads to a balanced workload distribution across training roles and maximizes hardware utilization.

Empirical observations indicate that generation time exhibits a unimodal relationship with respect to the resources allocated to a single component (e.g., the Sampler). To optimize resource placement, we employ the Ternary Search algorithm, as outlined in Algorithm \ref{alg:ts}. The algorithm is executed at regular intervals to ensure that the placement strategy remains optimal throughout the training process.

\begin{algorithm}[H]
  \label{alg:ts}
  \caption{Ternary Search for Dynamic Placement}
  \KwIn{Real generation time $f(x)$ with $x$ GPUs for sampler, number of GPUs $n$ }
  \KwOut{optimal $x^*$}
  $r = n, l = 1$ \\
  \While{$r - l > 0$}{
      $m_1 \leftarrow l + \frac{r - l}{3} , m_2 \leftarrow r - \frac{r - l}{3}$  \\
      \eIf{$f(m_1) > f(m_2)$}{
          $l \leftarrow m_1$
      }{
          $r \leftarrow m_2$
      }
  }
  $x^* \leftarrow l$ \\
  \Return{$x^*$}
  \end{algorithm}
\section{Implementation}

We implement WeChat-YATT using Python and PyTorch \cite{torch}. Our system combines vLLM \cite{vllm} and SGlang \cite{sglang} for generation serving with Megatron-Core \cite{mlm1, mlm2, mlm3, mlm4} as the training backend.

\subsection{Multi-processing}

No software is perfect; every piece of software inevitably has some flaws or limitations. Some programs may create conflicting global objects, or rely on global state that makes multi-tenancy impossible. These kinds of issue can lead to unexpected behavior, or difficulties when integrating different components. This can cause subtle or hard-to-detect issues when running multiple models in a single process. For example, some distributed training libraries use a global communication process group, making it impossible to assign different hybrid parallel plans to different models for optimal throughput. Due to the internal scheduling of enterprises, it is unrealistic to expect the open-source community to deliver bug fixes or undertake large-scale refactoring within a specific timeframe. This reality necessitates that enterprises plan ahead and develop their own solutions, ensuring that their projects can proceed according to their own schedules without being hindered by external dependencies.

We distribute all modules, whether for generation, training, or rewarding, across different processes, and enable them to collaborate via Remote Procedure Call. At the same time, we strive to minimize interference with the orchestration mechanisms used by these modules. By running each component in a separate process, we can avoid issues where multiple instances of certain components interfere with each other, which could otherwise impact throughput or introduce subtle bugs. Furthermore, by respecting the original design of each component and minimizing our interference with their orchestration, we allow the software to operate along well-tested code paths as much as possible. This approach not only improves stability and reliability, but also makes it easier to diagnose and resolve issues when they arise.

\subsection{Placement and RPC}

To keep the placement scheme as simple and predictable as possible (e.g., forming communication groups according to the GPU switch topology) and to minimize interference with component orchestration, we did not use Ray \cite{ray} for placement and RPC. Instead, we launch tasks via WeChat’s internal job scheduling system. Nevertheless, our code is designed to be loosely coupled, and we believe it can interoperate well with systems such as MPI \cite{openmpi} and SLURM \cite{slurm}. We do not doubt that Ray could also accomplish similar tasks.

To ensure RPC delivery and exactly-once semantics, we designed a simple and generic RPC mechanism: each RPC request is assigned a unique ID, and the result is cached on the server side until the client successfully retrieves it. The client then sends a request to clean up the cached RPC result. Since deep learning training systems typically only consider complete success (with all other cases treated as complete failure), error handling is greatly simplified. If the RPC returns an unexpected or undesired result, the controller simply terminates all processes.

Since it is impossible to completely avoid rare and untested code paths in components that may cause the system to hang, or low-probability scenarios within the internal cluster that may result in extremely slow CPU execution, we monitor the training progress to ensure it meets expectations. If the training progress falls below the expected threshold, the job is terminated, resources are reallocated, and the job is restarted.

\subsection{Idle Resources and Checkpointing}

Most WeChat services exhibit strong temporal fluctuations in resource usage, resulting in a large amount of idle resources during off-peak hours. We leverage these idle resources for non-urgent model training tasks.

To minimize training progress loss due to interruptions, we employ asynchronous checkpointing to improve checkpoint efficiency and increase checkpoint frequency. Additionally, when online services detect increased load and request resources from the training cluster, we attempt to save an on-demand checkpoint. If the checkpoint cannot be completed within the specified time, we abandon the current progress and release resources to prioritize online service needs.

To accommodate elastic resource scaling, we utilize distributed checkpointing and design the dataloader consumption state such that checkpoints can be reused across GPU clusters of varying sizes.

\subsection{Workload Balancing}

As noted in \citet{wlbllm} and \citet{megascale}, workload imbalance can create significant bubbles during training, becoming a major bottleneck. When sequence lengths are long, the main training bottleneck is attention computation, whose complexity is $s^2$ for a sequence of length $s$, while the cost of packing is much lower. For example, if a sequence is formed by concatenating two sequences of length $\frac s 2$, the total computation is $\frac {s^2} 2$, leading to severe load imbalance and device underutilization. This issue is even more pronounced with post-training data, which often varies greatly in length.

Compared to complex combinatorial optimization approaches, we find a much simpler solution for workload balancing: instead of sequence packing, we simply sort data by simulated workload ($s^2$), greatly simplifying the process and even improving computational efficiency.

Most of the time, the length distribution of the data is normal. This means that, in practice, the workload is also normally distributed, which makes it easy for us to construct global batches with very little padding. In the worst case, if the data lengths are randomly distributed, the computational waste is still minimal. Suppose the maximum length is $s$,  and the local reordering batch buffer size is $B$, then the useless computation within a single batch does not exceed $1 - \left( \frac{B-1}{B} \right)^2$. For example, when $s=1M, B=16$, at most 8.6\% of the computation is wasted.

Naively sorting data by workload(length) may introduce distribution bias. To address this, we first bucket data according to the global batch size, then shuffle the buckets to ensure data is randomly distributed. Our experiments show that this method has almost no impact on model accuracy.

\subsection{Distributed Attention}

Currently, most implementations achieve context parallelism using ring attention for both causal and full attention. To support more complex attention masks (e.g., Gemma-3 \cite{gemma3}), inspired by Llama-3 \cite{llama3}, we implement distributed attention based on CCL all-gather, where we first all-gather the key (K) and value (V) tensors, and then compute attention output for the local query (Q) tensor chunk.

To reduce the memory footprint of all-gathered key-value pairs, inspired by open source community contributors, we process only a subset of attention heads at a time and overlap KV communication with attention computation. This makes it feasible to train sequences up to 1 million tokens in length.

\subsection{Train Data Storage}

Multimodal data is often extremely large, both in quantity and size \cite{qwen25vl, gemma3}. While algorithm developers typically use JSONL files for flexible storage of text data, images and videos cannot be easily stored in this format.

Storing massive numbers of images directly in a distributed file system can easily exceed file number quota, putting pressure on the distributed file system name nodes. To address this, we adopt private key-value storage solutions (e.g. FeatureKV and UnionDB), on top of our private service discovery and distributed file system (Wechat File System) \cite{wxpaxos}, to serve training data. By leveraging more efficient storage engine data structures, we achieve more efficient and convenient storage and access for training data.

\section{Evaluation}

We conduct our experiments on a cluster of NVIDIA GPUs, where each node is equipped with eight NVIDIA H20-96GB GPUs interconnected via NVLink. The inter-node communication leverages RDMA with a bandwidth of 200 Gbps. The software stack comprises CUDA 12.4, PyTorch 2.6, Megatron-core 0.12.2, and SGLang 0.4.6.post5. While the evaluation presented in this paper utilizes a relatively small number of GPUs to validate WeChat-YATT, it is noteworthy that large-scale GPU clusters—comprising up to thousands of GPUs—are deployed in WeChat’s internal production environment. However, due to compliance constraints, further details regarding the production deployment cannot be disclosed.

\subsection{Generative Reward Model as Critic in Training}

We begin by evaluating WeChat-YATT in conjunction with a generative reward model on a mathematical reasoning task. Specifically, we employ the Qwen 2.5 model series \cite{qwen25math} as the backbone models and utilize the GSM8K dataset \cite{gsm8k} for training.

We conduct experiments with the following methods:
\begin{itemize}
  \item \textbf{VeRL\cite{verl}:} The framework is an efficient reinforcement learning platform designed to support scalable, distributed training and evaluation of RL algorithms across diverse environments proposed be seed team.
  \item \textbf{Full Colocated WeChat-YATT:} This framework is implemented over WeChat-YATT in a full colocate manner. All components in the framework are colocated across clusters sharing all resources.
  
  \item \textbf{Partial Colocated WeChat-YATT:} In this framework, a partial colocation strategy is employed to facilitate dynamic sampling.The actor is colocated with other components, while the sampler—responsible for actor rollouts—operates independently alongside the generative reward model. 
\end{itemize}

The metric of the evalutaion is execution time in unit of seconds to reflect the overall throughput of different framework.

The comparative evaluation of the \textit{Generative Reward Model as Critic in Training} is conducted in two parts. First, we assess the throughput of the generative reward model under a full colocated WeChat-YATT deployment compared with sota framework VeRL. Subsequently, we investigate the impact of dynamic sampling in specific scenarios, where frequent swapping introduces additional overhead and exacerbates long-tail latency.

\subsubsection{Co-locate Generative Reward Model}

In the full colocated WeChat-YATT setup, we conduct experiments on the generative reward model compared with VeRL under two distinct scenarios. The first scenario features a balanced parameter size between the generative reward model and the actor model, while the second scenario considers a highly unbalanced configuration. As VeRL currently supports only external calls to the generative reward model, we allocate computational resources in proportion to the parameter size ratio between the generative reward and actor models.

For the first comparation scenario setup, we select the VeRL Generative Reward demo. In this configurationof VeRL, GenRM-CI-Test-1.5B serves as the external reward model, utilizing 8 GPUs for inference. SGlang is employed as the inference engine, with a data parallelism size (dp-size) of 8. An additional 8 GPUs are allocated to deploy the actor model, Qwen2.5-3B-Instruct, with a tensor parallelism size (tp-size) of 2. For the training backend, we adopt the state-of-the-art Megatron-LM framework. To ensure a fair comparison, the WeChat-YATT setup mirrors the VeRL configuration, employing a full colocated deployment with 16 GPUs.

\begin{figure}[htbp]
  \centering
  \begin{subfigure}[b]{0.55\textwidth}
      \centering
      \includegraphics[width=\textwidth]{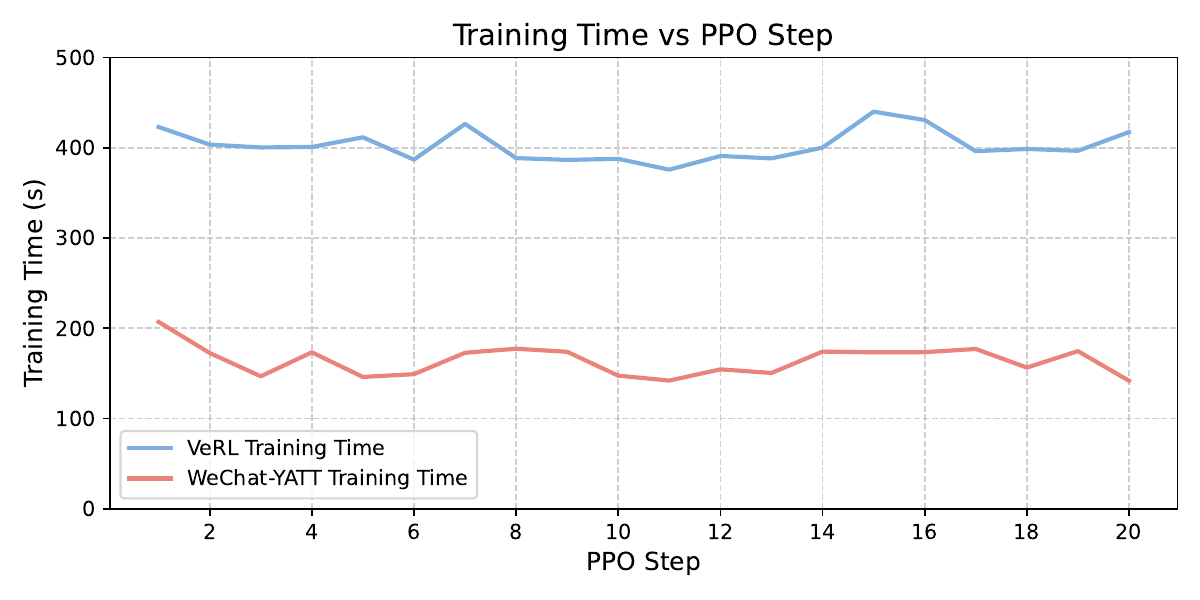}
      \caption{Training Time of Full Colocated WeChat-YATT and VeRL}
      \label{fig:sc1-tt}
  \end{subfigure}
  \hspace{0.05\textwidth}
  \begin{subfigure}[b]{0.35\textwidth}
      \centering
      \includegraphics[width=\textwidth]{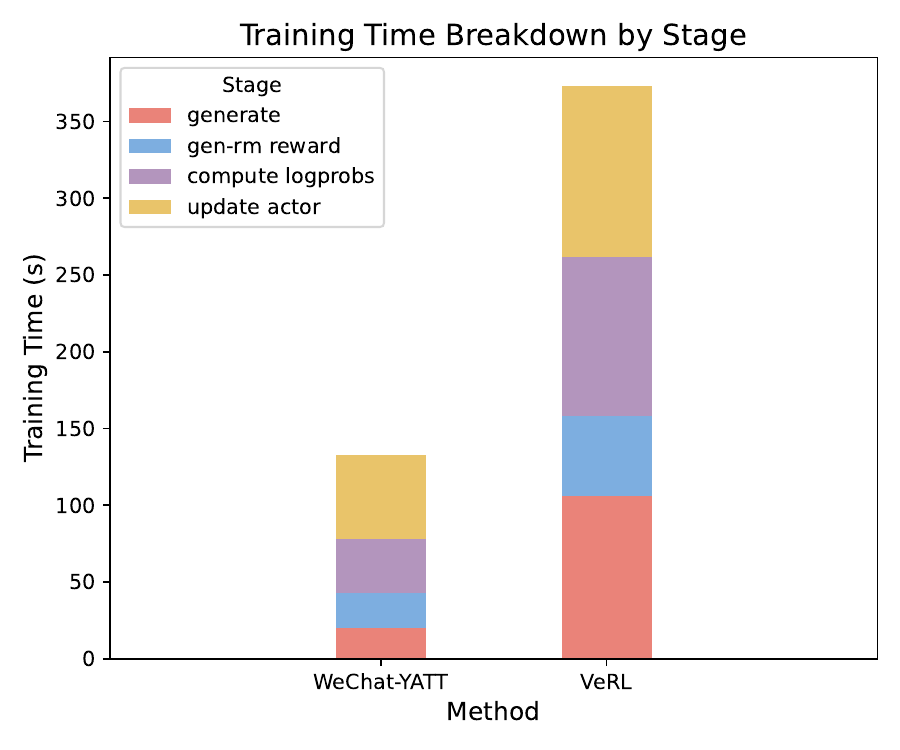}
      \caption{Time consumption breakdown}
      \label{fig:sc1-tcb}
  \end{subfigure}
  \caption{Training time and time consumption breakdown with GenRM-CI-Test-1.5B as gen-rm, and Qwen2.5-3B-Instruct as actor}
  \label{fig:tt}
\end{figure}

As illustrated in Figure \ref{fig:tt}, WeChat-YATT achieves a 59.2\% reduction in average training time compared to VeRL. This improvement is primarily attributed to WeChat-YATT’s full colocated strategy, which enables the utilization of all available GPUs throughout the training process. Specifically, WeChat-YATT offloads the parameters of the inference engine during training and reloads them as needed, thereby minimizing idle periods for both the generative reward model and the actor model. To further investigate the performance advantage of WeChat-YATT in this scenario, we analyze four key stages of a PPO step. As shown in Figure \ref{fig:sc1-tcb}, each stage benefits from the partial colocated strategy, as all 16 GPUs are actively engaged due to the offload-onload mechanism, resulting in consistently faster execution across all stages.

\begin{figure}[htbp]
  \centering
  \begin{subfigure}[b]{0.55\textwidth}
      \centering
      \includegraphics[width=\textwidth]{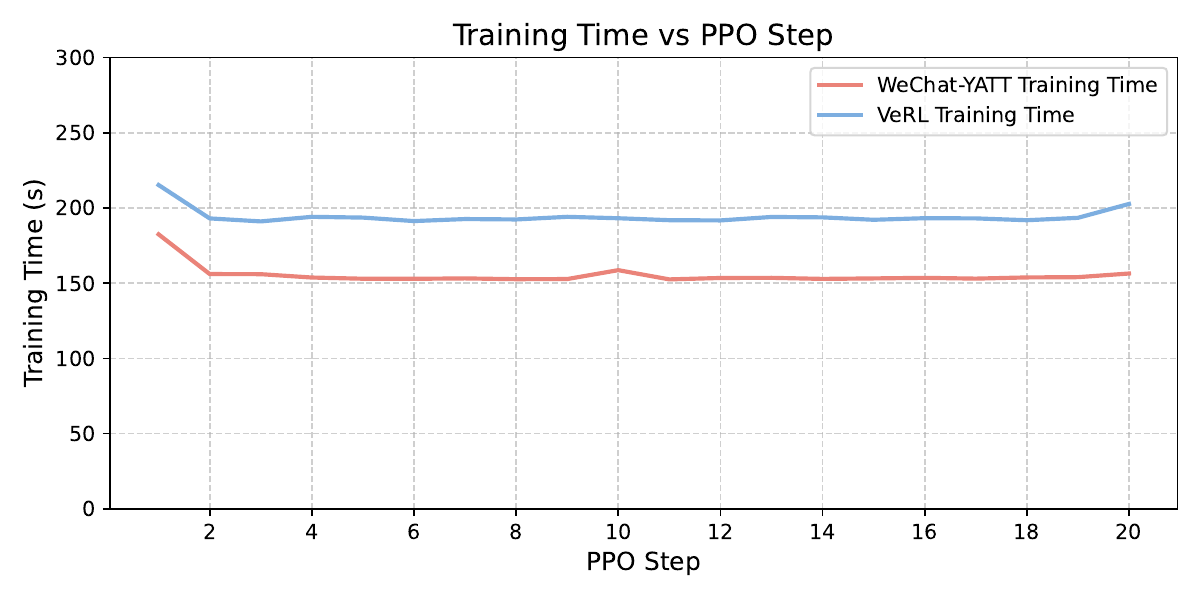}
      \caption{Training Time of Full Colocated WeChat-YATT and VeRL}
      \label{fig:sc2-tt}
  \end{subfigure}
  \hspace{0.05\textwidth}
  \begin{subfigure}[b]{0.35\textwidth}
      \centering
      \includegraphics[width=\textwidth]{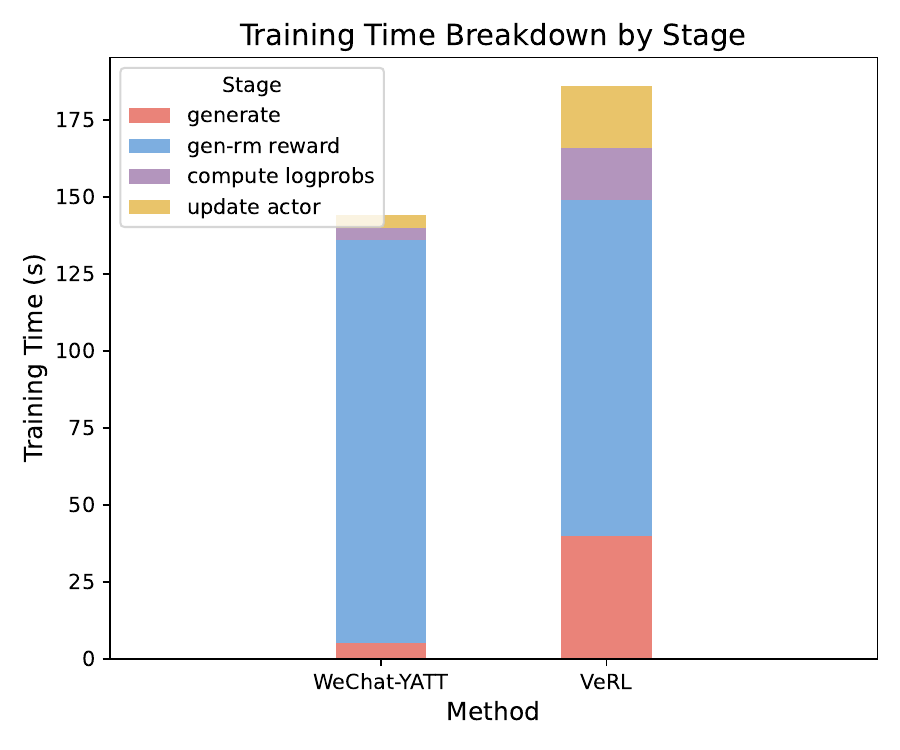}
      \caption{Time consumption breakdown}
      \label{fig:sc2-tcb}
  \end{subfigure}
  \caption{Training time and time consumption breakdown with Qwen2.5-Math-72B-Instruct as gen-rm, and Qwen2.5-Math-1.5B as actor}
  \label{fig:tt2}
\end{figure}

We further evaluate performance in a scenario involving a large generative reward model, Qwen2.5-Math-72B-Instruct, paired with a smaller actor model, Qwen2.5-Math-1.5B. This configuration corresponds to the generative reward model demonstration provided in our open-source codebase. For the VeRL setup, 24 GPUs are allocated for inference, while an additional 8 GPUs are dedicated to actor model rollout generation and training. In contrast, WeChat-YATT employ a full colocated configuration utilizing 32 GPUs, leveraging an offload/onload strategy. Megatron-LM is used as the training backend in both cases.

As depicted in Figure \ref{fig:tt2}, WeChat-YATT achieves a 20.0\% reduction in average training time across PPO training steps, as detailed in Figure \ref{fig:sc2-tt}. In this extreme scenario, the primary bottleneck is the generative reward model evaluation, as shown in Figure \ref{fig:sc2-tcb}. While WeChat-YATT optimizes the actor rollout generation and training phases, this comes at the cost of additional overhead associated with offloading and onloading the large generative reward model. Notably, even though WeChat-YATT utilizes 32 GPUs for the generative reward model—compared to 24 GPUs in the VeRL setup—it exhibits a slightly longer reward evaluation time. This is attributable to the dominance of onload-offload overhead in scenarios characterized by low sequence length and small rollout batch size, where the additional communication and memory management costs outweigh the benefits of increased GPU parallelism. Nevertheless, WeChat-YATT accelerates the remaining stages of the training pipeline, resulting in a lower overall training time compared to VeRL.

\begin{table}[htbp]
  \centering
  \caption{Rollout Time Usage under Different Sample Rejected Rates}
  \begin{tabular}{lcccccccc}
  \toprule
  \multirow{2}{*}{} & \multicolumn{2}{c}{10\%} & \multicolumn{2}{c}{20\%} & \multicolumn{2}{c}{30\%} & \multicolumn{2}{c}{40\%} \\
  \cmidrule(lr){2-3} \cmidrule(lr){4-5} \cmidrule(lr){6-7} \cmidrule(lr){8-9}
                    & Total (s) & Avg. (s) & Total (s) & Avg. (s) & Total (s) & Avg. (s) & Total (s) & Avg. (s) \\
  \midrule
  Full Colocate     & 1987.1    & 39.7     & 2350.6    & 47.0     & 2838.4         & 56.8        & 3457.3         & 69.1        \\
  Partial Colocate  & 1509.1    & 30.1     & 1968.5    & 39.4     & 2315.9         & 46.3        & 2684.4         & 53.7        \\
  \bottomrule
  \label{tab:reject_rates}
  \end{tabular}
  \end{table}

\subsubsection{Dynamic Sampling \label{subsec:ds}}

To assess the performance of the partial colocated WeChat-YATT framework in a dynamic sampling context, we conduct experiments in which varying sample rejection rates are applied during the rollout phase. Consistent with previous experimental settings, GenRM-CI-Test-1.5B is employed as the generative reward model, while Qwen2.5-3B-Instruct served as the actor model. The sample rejection rates are varied from 10\% to 40\%, and the framework need to re-sample rejected rollouts until all samples are accepted. The primary metric of interest is the time consumption observed during the rollout phase under these different rejection rates.

As shown in Table \ref{tab:reject_rates}, the rollout time for both frameworks increases as the sample rejection rate rises. However, the partial colocated WeChat-YATT consistently demonstrates lower rollout time across all rejection rates when compared to the full colocated approach. Specifically, the partial colocated framework achieves a reduction in rollout time usage ranging from 16.2\% to 24.1\% relative to the full colocated configuration.

This performance improvement can be attributed to the asynchronous interaction between the actor rollouts and the generative reward model in the partial colocated design. In this setup, each completed microbatch of actor rollouts is immediately dispatched to the standalone generative reward model in an asynchronous manner. This process effectively mitigates the long-tail latency that often constrains throughput in the full colocated WeChat-YATT framework. Furthermore, the frequent context switching between actor rollouts and the generative model in the full colocated setup introduces additional overhead, which is substantially reduced in the partial colocated architecture.

\subsection{Dynamic Placement}

\begin{figure}
  \centering
  \includegraphics[width=.85\linewidth]{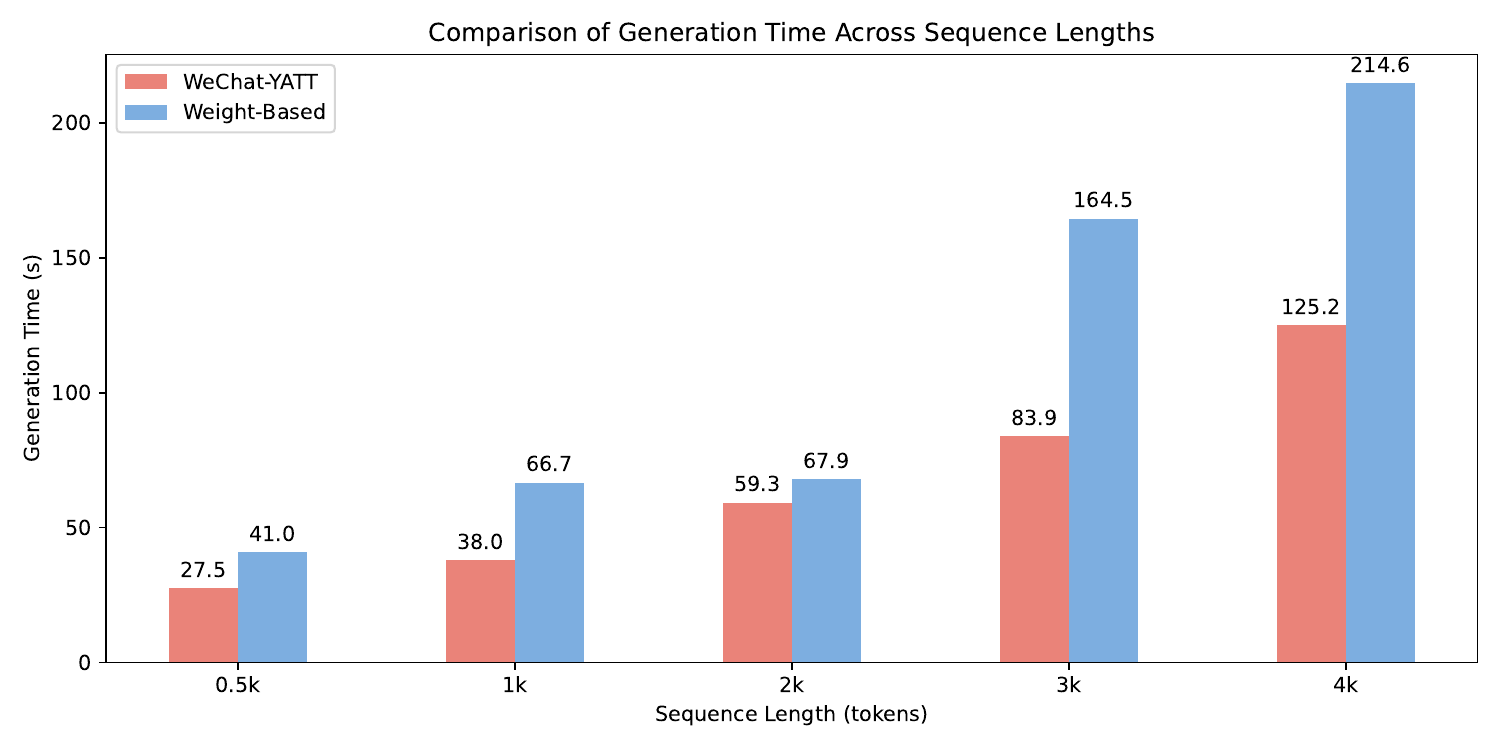}
  \caption{Dynamic placement algorithm generation time compared with Weighted-Based placement}
\label{fig:dp_exp}
\end{figure}

We adopt the same experimental setup as described in subsection \ref{subsec:ds}. In this set of experiments, we vary the generation sequence length to examine how generation time changes as the sequence length increases. For comparison, we employ the Weighted-Based placement strategy, in which computational resources are allocated in proportion to the model weights of the generative reward model and the actor model.

The experiments results are presented in Figure \ref{fig:dp_exp}. The reported generation time metric represents the average across all PPO steps. As expected, our results demonstrate that WeChat-YATT consistently achieves lower generation times than the weighted-based resource placement strategy across different generation sequence lengths.

The superior performance of WeChat-YATT can be attributed to its ability to identify the optimal resource placement strategy during training via the ternary search algorithm. Furthermore, by partially colocating both the policy and generative reward models in GPU memory, WeChat-YATT effectively eliminates swapping overhead. This approach also substantially mitigates the long-tail effect, further enhancing system efficiency.

\section{Conclusion}

In this work, we present WeChat-YATT, a simple, scalable, and balanced RLHF training framework that addresses critical bottlenecks in large-scale, multi-model reinforcement learning workflows. By introducing a parallel controller programming model and a dynamic scaling placement schema, WeChat-YATT effectively mitigates memory and communication bottlenecks, and adapts to dynamic workloads with improved hardware utilization. Our design emphasizes practicality and flexibility, aiming to reduce engineering complexity while supporting efficient orchestration of complex RLHF pipelines.

WeChat-YATT has already produced various models that deployed in WeChat’s production environment, supporting real-world business applications that serve the entire user base. This practical deployment demonstrates the framework’s robustness and scalability. Although challenges remain in optimizing resource allocation and accommodating increasingly diverse model architectures, we believe WeChat-YATT offers a strong foundation for advancing research and deploying large-scale, human-aligned models in practice.

\bibliographystyle{unsrt}  
\bibliography{refs}

\begin{thebibliography}{59}
\providecommand{\natexlab}[1]{#1}
\providecommand{\url}[1]{\texttt{#1}}
\expandafter\ifx\csname urlstyle\endcsname\relax
  \providecommand{\doi}[1]{doi: #1}\else
  \providecommand{\doi}{doi: \begingroup \urlstyle{rm}\Url}\fi

\bibitem[OpenAI et~al.(2024)OpenAI, Achiam, Adler, Agarwal, Ahmad, Akkaya, Aleman, Almeida, Altenschmidt, Altman, Anadkat, Avila, Babuschkin, Balaji, Balcom, Baltescu, Bao, Bavarian, Belgum, Bello, Berdine, Bernadett-Shapiro, Berner, Bogdonoff, Boiko, Boyd, Brakman, Brockman, Brooks, Brundage, Button, Cai, Campbell, Cann, Carey, Carlson, Carmichael, Chan, Chang, Chantzis, Chen, Chen, Chen, Chen, Chen, Chess, Cho, Chu, Chung, Cummings, Currier, Dai, Decareaux, Degry, Deutsch, Deville, Dhar, Dohan, Dowling, Dunning, Ecoffet, Eleti, Eloundou, Farhi, Fedus, Felix, Fishman, Forte, Fulford, Gao, Georges, Gibson, Goel, Gogineni, Goh, Gontijo-Lopes, Gordon, Grafstein, Gray, Greene, Gross, Gu, Guo, Hallacy, Han, Harris, He, Heaton, Heidecke, Hesse, Hickey, Hickey, Hoeschele, Houghton, Hsu, Hu, Hu, Huizinga, Jain, Jain, Jang, Jiang, Jiang, Jin, Jin, Jomoto, Jonn, Jun, Kaftan, Łukasz Kaiser, Kamali, Kanitscheider, Keskar, Khan, Kilpatrick, Kim, Kim, Kim, Kirchner, Kiros, Knight, Kokotajlo, Łukasz Kondraciuk, Kondrich, Konstantinidis, Kosic, Krueger, Kuo, Lampe, Lan, Lee, Leike, Leung, Levy, Li, Lim, Lin, Lin, Litwin, Lopez, Lowe, Lue, Makanju, Malfacini, Manning, Markov, Markovski, Martin, Mayer, Mayne, McGrew, McKinney, McLeavey, McMillan, McNeil, Medina, Mehta, Menick, Metz, Mishchenko, Mishkin, Monaco, Morikawa, Mossing, Mu, Murati, Murk, Mély, Nair, Nakano, Nayak, Neelakantan, Ngo, Noh, Ouyang, O'Keefe, Pachocki, Paino, Palermo, Pantuliano, Parascandolo, Parish, Parparita, Passos, Pavlov, Peng, Perelman, de~Avila Belbute~Peres, Petrov, de~Oliveira~Pinto, Michael, Pokorny, Pokrass, Pong, Powell, Power, Power, Proehl, Puri, Radford, Rae, Ramesh, Raymond, Real, Rimbach, Ross, Rotsted, Roussez, Ryder, Saltarelli, Sanders, Santurkar, Sastry, Schmidt, Schnurr, Schulman, Selsam, Sheppard, Sherbakov, Shieh, Shoker, Shyam, Sidor, Sigler, Simens, Sitkin, Slama, Sohl, Sokolowsky, Song, Staudacher, Such, Summers, Sutskever, Tang, Tezak, Thompson, Tillet, Tootoonchian, Tseng, Tuggle, Turley, Tworek, Uribe, Vallone, Vijayvergiya, Voss, Wainwright, Wang, Wang, Wang, Ward, Wei, Weinmann, Welihinda, Welinder, Weng, Weng, Wiethoff, Willner, Winter, Wolrich, Wong, Workman, Wu, Wu, Wu, Xiao, Xu, Yoo, Yu, Yuan, Zaremba, Zellers, Zhang, Zhang, Zhao, Zheng, Zhuang, Zhuk, and Zoph]{gpt4}
OpenAI, Josh Achiam, Steven Adler, Sandhini Agarwal, Lama Ahmad, Ilge Akkaya, Florencia~Leoni Aleman, Diogo Almeida, Janko Altenschmidt, Sam Altman, Shyamal Anadkat, Red Avila, Igor Babuschkin, Suchir Balaji, Valerie Balcom, Paul Baltescu, Haiming Bao, Mohammad Bavarian, Jeff Belgum, Irwan Bello, Jake Berdine, Gabriel Bernadett-Shapiro, Christopher Berner, Lenny Bogdonoff, Oleg Boiko, Madelaine Boyd, Anna-Luisa Brakman, Greg Brockman, Tim Brooks, Miles Brundage, Kevin Button, Trevor Cai, Rosie Campbell, Andrew Cann, Brittany Carey, Chelsea Carlson, Rory Carmichael, Brooke Chan, Che Chang, Fotis Chantzis, Derek Chen, Sully Chen, Ruby Chen, Jason Chen, Mark Chen, Ben Chess, Chester Cho, Casey Chu, Hyung~Won Chung, Dave Cummings, Jeremiah Currier, Yunxing Dai, Cory Decareaux, Thomas Degry, Noah Deutsch, Damien Deville, Arka Dhar, David Dohan, Steve Dowling, Sheila Dunning, Adrien Ecoffet, Atty Eleti, Tyna Eloundou, David Farhi, Liam Fedus, Niko Felix, Simón~Posada Fishman, Juston Forte, Isabella Fulford, Leo Gao, Elie Georges, Christian Gibson, Vik Goel, Tarun Gogineni, Gabriel Goh, Rapha Gontijo-Lopes, Jonathan Gordon, Morgan Grafstein, Scott Gray, Ryan Greene, Joshua Gross, Shixiang~Shane Gu, Yufei Guo, Chris Hallacy, Jesse Han, Jeff Harris, Yuchen He, Mike Heaton, Johannes Heidecke, Chris Hesse, Alan Hickey, Wade Hickey, Peter Hoeschele, Brandon Houghton, Kenny Hsu, Shengli Hu, Xin Hu, Joost Huizinga, Shantanu Jain, Shawn Jain, Joanne Jang, Angela Jiang, Roger Jiang, Haozhun Jin, Denny Jin, Shino Jomoto, Billie Jonn, Heewoo Jun, Tomer Kaftan, Łukasz Kaiser, Ali Kamali, Ingmar Kanitscheider, Nitish~Shirish Keskar, Tabarak Khan, Logan Kilpatrick, Jong~Wook Kim, Christina Kim, Yongjik Kim, Jan~Hendrik Kirchner, Jamie Kiros, Matt Knight, Daniel Kokotajlo, Łukasz Kondraciuk, Andrew Kondrich, Aris Konstantinidis, Kyle Kosic, Gretchen Krueger, Vishal Kuo, Michael Lampe, Ikai Lan, Teddy Lee, Jan Leike, Jade Leung, Daniel Levy, Chak~Ming Li, Rachel Lim, Molly Lin, Stephanie Lin, Mateusz Litwin, Theresa Lopez, Ryan Lowe, Patricia Lue, Anna Makanju, Kim Malfacini, Sam Manning, Todor Markov, Yaniv Markovski, Bianca Martin, Katie Mayer, Andrew Mayne, Bob McGrew, Scott~Mayer McKinney, Christine McLeavey, Paul McMillan, Jake McNeil, David Medina, Aalok Mehta, Jacob Menick, Luke Metz, Andrey Mishchenko, Pamela Mishkin, Vinnie Monaco, Evan Morikawa, Daniel Mossing, Tong Mu, Mira Murati, Oleg Murk, David Mély, Ashvin Nair, Reiichiro Nakano, Rajeev Nayak, Arvind Neelakantan, Richard Ngo, Hyeonwoo Noh, Long Ouyang, Cullen O'Keefe, Jakub Pachocki, Alex Paino, Joe Palermo, Ashley Pantuliano, Giambattista Parascandolo, Joel Parish, Emy Parparita, Alex Passos, Mikhail Pavlov, Andrew Peng, Adam Perelman, Filipe de~Avila Belbute~Peres, Michael Petrov, Henrique~Ponde de~Oliveira~Pinto, Michael, Pokorny, Michelle Pokrass, Vitchyr~H. Pong, Tolly Powell, Alethea Power, Boris Power, Elizabeth Proehl, Raul Puri, Alec Radford, Jack Rae, Aditya Ramesh, Cameron Raymond, Francis Real, Kendra Rimbach, Carl Ross, Bob Rotsted, Henri Roussez, Nick Ryder, Mario Saltarelli, Ted Sanders, Shibani Santurkar, Girish Sastry, Heather Schmidt, David Schnurr, John Schulman, Daniel Selsam, Kyla Sheppard, Toki Sherbakov, Jessica Shieh, Sarah Shoker, Pranav Shyam, Szymon Sidor, Eric Sigler, Maddie Simens, Jordan Sitkin, Katarina Slama, Ian Sohl, Benjamin Sokolowsky, Yang Song, Natalie Staudacher, Felipe~Petroski Such, Natalie Summers, Ilya Sutskever, Jie Tang, Nikolas Tezak, Madeleine~B. Thompson, Phil Tillet, Amin Tootoonchian, Elizabeth Tseng, Preston Tuggle, Nick Turley, Jerry Tworek, Juan Felipe~Cerón Uribe, Andrea Vallone, Arun Vijayvergiya, Chelsea Voss, Carroll Wainwright, Justin~Jay Wang, Alvin Wang, Ben Wang, Jonathan Ward, Jason Wei, CJ~Weinmann, Akila Welihinda, Peter Welinder, Jiayi Weng, Lilian Weng, Matt Wiethoff, Dave Willner, Clemens Winter, Samuel Wolrich, Hannah Wong, Lauren Workman, Sherwin Wu, Jeff Wu, Michael Wu, Kai Xiao, Tao Xu, Sarah Yoo, Kevin Yu, Qiming Yuan, Wojciech Zaremba, Rowan Zellers, Chong Zhang, Marvin Zhang, Shengjia Zhao, Tianhao Zheng, Juntang Zhuang, William Zhuk, and Barret Zoph.
\newblock Gpt-4 technical report, 2024.
\newblock URL \url{https://arxiv.org/abs/2303.08774}.

\bibitem[Grattafiori et~al.(2024)Grattafiori, Dubey, Jauhri, Pandey, Kadian, Al-Dahle, Letman, Mathur, Schelten, Vaughan, Yang, Fan, Goyal, Hartshorn, Yang, Mitra, Sravankumar, Korenev, Hinsvark, Rao, Zhang, Rodriguez, Gregerson, Spataru, Roziere, Biron, Tang, Chern, Caucheteux, Nayak, Bi, Marra, McConnell, Keller, Touret, Wu, Wong, Ferrer, Nikolaidis, Allonsius, Song, Pintz, Livshits, Wyatt, Esiobu, Choudhary, Mahajan, Garcia-Olano, Perino, Hupkes, Lakomkin, AlBadawy, Lobanova, Dinan, Smith, Radenovic, Guzmán, Zhang, Synnaeve, Lee, Anderson, Thattai, Nail, Mialon, Pang, Cucurell, Nguyen, Korevaar, Xu, Touvron, Zarov, Ibarra, Kloumann, Misra, Evtimov, Zhang, Copet, Lee, Geffert, Vranes, Park, Mahadeokar, Shah, van~der Linde, Billock, Hong, Lee, Fu, Chi, Huang, Liu, Wang, Yu, Bitton, Spisak, Park, Rocca, Johnstun, Saxe, Jia, Alwala, Prasad, Upasani, Plawiak, Li, Heafield, Stone, El-Arini, Iyer, Malik, Chiu, Bhalla, Lakhotia, Rantala-Yeary, van~der Maaten, Chen, Tan, Jenkins, Martin, Madaan, Malo, Blecher, Landzaat, de~Oliveira, Muzzi, Pasupuleti, Singh, Paluri, Kardas, Tsimpoukelli, Oldham, Rita, Pavlova, Kambadur, Lewis, Si, Singh, Hassan, Goyal, Torabi, Bashlykov, Bogoychev, Chatterji, Zhang, Duchenne, Çelebi, Alrassy, Zhang, Li, Vasic, Weng, Bhargava, Dubal, Krishnan, Koura, Xu, He, Dong, Srinivasan, Ganapathy, Calderer, Cabral, Stojnic, Raileanu, Maheswari, Girdhar, Patel, Sauvestre, Polidoro, Sumbaly, Taylor, Silva, Hou, Wang, Hosseini, Chennabasappa, Singh, Bell, Kim, Edunov, Nie, Narang, Raparthy, Shen, Wan, Bhosale, Zhang, Vandenhende, Batra, Whitman, Sootla, Collot, Gururangan, Borodinsky, Herman, Fowler, Sheasha, Georgiou, Scialom, Speckbacher, Mihaylov, Xiao, Karn, Goswami, Gupta, Ramanathan, Kerkez, Gonguet, Do, Vogeti, Albiero, Petrovic, Chu, Xiong, Fu, Meers, Martinet, Wang, Wang, Tan, Xia, Xie, Jia, Wang, Goldschlag, Gaur, Babaei, Wen, Song, Zhang, Li, Mao, Coudert, Yan, Chen, Papakipos, Singh, Srivastava, Jain, Kelsey, Shajnfeld, Gangidi, Victoria, Goldstand, Menon, Sharma, Boesenberg, Baevski, Feinstein, Kallet, Sangani, Teo, Yunus, Lupu, Alvarado, Caples, Gu, Ho, Poulton, Ryan, Ramchandani, Dong, Franco, Goyal, Saraf, Chowdhury, Gabriel, Bharambe, Eisenman, Yazdan, James, Maurer, Leonhardi, Huang, Loyd, Paola, Paranjape, Liu, Wu, Ni, Hancock, Wasti, Spence, Stojkovic, Gamido, Montalvo, Parker, Burton, Mejia, Liu, Wang, Kim, Zhou, Hu, Chu, Cai, Tindal, Feichtenhofer, Gao, Civin, Beaty, Kreymer, Li, Adkins, Xu, Testuggine, David, Parikh, Liskovich, Foss, Wang, Le, Holland, Dowling, Jamil, Montgomery, Presani, Hahn, Wood, Le, Brinkman, Arcaute, Dunbar, Smothers, Sun, Kreuk, Tian, Kokkinos, Ozgenel, Caggioni, Kanayet, Seide, Florez, Schwarz, Badeer, Swee, Halpern, Herman, Sizov, Guangyi, Zhang, Lakshminarayanan, Inan, Shojanazeri, Zou, Wang, Zha, Habeeb, Rudolph, Suk, Aspegren, Goldman, Zhan, Damlaj, Molybog, Tufanov, Leontiadis, Veliche, Gat, Weissman, Geboski, Kohli, Lam, Asher, Gaya, Marcus, Tang, Chan, Zhen, Reizenstein, Teboul, Zhong, Jin, Yang, Cummings, Carvill, Shepard, McPhie, Torres, Ginsburg, Wang, Wu, U, Saxena, Khandelwal, Zand, Matosich, Veeraraghavan, Michelena, Li, Jagadeesh, Huang, Chawla, Huang, Chen, Garg, A, Silva, Bell, Zhang, Guo, Yu, Moshkovich, Wehrstedt, Khabsa, Avalani, Bhatt, Mankus, Hasson, Lennie, Reso, Groshev, Naumov, Lathi, Keneally, Liu, Seltzer, Valko, Restrepo, Patel, Vyatskov, Samvelyan, Clark, Macey, Wang, Hermoso, Metanat, Rastegari, Bansal, Santhanam, Parks, White, Bawa, Singhal, Egebo, Usunier, Mehta, Laptev, Dong, Cheng, Chernoguz, Hart, Salpekar, Kalinli, Kent, Parekh, Saab, Balaji, Rittner, Bontrager, Roux, Dollar, Zvyagina, Ratanchandani, Yuvraj, Liang, Alao, Rodriguez, Ayub, Murthy, Nayani, Mitra, Parthasarathy, Li, Hogan, Battey, Wang, Howes, Rinott, Mehta, Siby, Bondu, Datta, Chugh, Hunt, Dhillon, Sidorov, Pan, Mahajan, Verma, Yamamoto, Ramaswamy, Lindsay, Lindsay, Feng, Lin, Zha, Patil, Shankar, Zhang, Zhang, Wang, Agarwal, Sajuyigbe, Chintala, Max, Chen, Kehoe, Satterfield, Govindaprasad, Gupta, Deng, Cho, Virk, Subramanian, Choudhury, Goldman, Remez, Glaser, Best, Koehler, Robinson, Li, Zhang, Matthews, Chou, Shaked, Vontimitta, Ajayi, Montanez, Mohan, Kumar, Mangla, Ionescu, Poenaru, Mihailescu, Ivanov, Li, Wang, Jiang, Bouaziz, Constable, Tang, Wu, Wang, Wu, Gao, Kleinman, Chen, Hu, Jia, Qi, Li, Zhang, Zhang, Adi, Nam, Yu, Wang, Zhao, Hao, Qian, Li, He, Rait, DeVito, Rosnbrick, Wen, Yang, Zhao, and Ma]{llama3}
Aaron Grattafiori, Abhimanyu Dubey, Abhinav Jauhri, Abhinav Pandey, Abhishek Kadian, Ahmad Al-Dahle, Aiesha Letman, Akhil Mathur, Alan Schelten, Alex Vaughan, Amy Yang, Angela Fan, Anirudh Goyal, Anthony Hartshorn, Aobo Yang, Archi Mitra, Archie Sravankumar, Artem Korenev, Arthur Hinsvark, Arun Rao, Aston Zhang, Aurelien Rodriguez, Austen Gregerson, Ava Spataru, Baptiste Roziere, Bethany Biron, Binh Tang, Bobbie Chern, Charlotte Caucheteux, Chaya Nayak, Chloe Bi, Chris Marra, Chris McConnell, Christian Keller, Christophe Touret, Chunyang Wu, Corinne Wong, Cristian~Canton Ferrer, Cyrus Nikolaidis, Damien Allonsius, Daniel Song, Danielle Pintz, Danny Livshits, Danny Wyatt, David Esiobu, Dhruv Choudhary, Dhruv Mahajan, Diego Garcia-Olano, Diego Perino, Dieuwke Hupkes, Egor Lakomkin, Ehab AlBadawy, Elina Lobanova, Emily Dinan, Eric~Michael Smith, Filip Radenovic, Francisco Guzmán, Frank Zhang, Gabriel Synnaeve, Gabrielle Lee, Georgia~Lewis Anderson, Govind Thattai, Graeme Nail, Gregoire Mialon, Guan Pang, Guillem Cucurell, Hailey Nguyen, Hannah Korevaar, Hu~Xu, Hugo Touvron, Iliyan Zarov, Imanol~Arrieta Ibarra, Isabel Kloumann, Ishan Misra, Ivan Evtimov, Jack Zhang, Jade Copet, Jaewon Lee, Jan Geffert, Jana Vranes, Jason Park, Jay Mahadeokar, Jeet Shah, Jelmer van~der Linde, Jennifer Billock, Jenny Hong, Jenya Lee, Jeremy Fu, Jianfeng Chi, Jianyu Huang, Jiawen Liu, Jie Wang, Jiecao Yu, Joanna Bitton, Joe Spisak, Jongsoo Park, Joseph Rocca, Joshua Johnstun, Joshua Saxe, Junteng Jia, Kalyan~Vasuden Alwala, Karthik Prasad, Kartikeya Upasani, Kate Plawiak, Ke~Li, Kenneth Heafield, Kevin Stone, Khalid El-Arini, Krithika Iyer, Kshitiz Malik, Kuenley Chiu, Kunal Bhalla, Kushal Lakhotia, Lauren Rantala-Yeary, Laurens van~der Maaten, Lawrence Chen, Liang Tan, Liz Jenkins, Louis Martin, Lovish Madaan, Lubo Malo, Lukas Blecher, Lukas Landzaat, Luke de~Oliveira, Madeline Muzzi, Mahesh Pasupuleti, Mannat Singh, Manohar Paluri, Marcin Kardas, Maria Tsimpoukelli, Mathew Oldham, Mathieu Rita, Maya Pavlova, Melanie Kambadur, Mike Lewis, Min Si, Mitesh~Kumar Singh, Mona Hassan, Naman Goyal, Narjes Torabi, Nikolay Bashlykov, Nikolay Bogoychev, Niladri Chatterji, Ning Zhang, Olivier Duchenne, Onur Çelebi, Patrick Alrassy, Pengchuan Zhang, Pengwei Li, Petar Vasic, Peter Weng, Prajjwal Bhargava, Pratik Dubal, Praveen Krishnan, Punit~Singh Koura, Puxin Xu, Qing He, Qingxiao Dong, Ragavan Srinivasan, Raj Ganapathy, Ramon Calderer, Ricardo~Silveira Cabral, Robert Stojnic, Roberta Raileanu, Rohan Maheswari, Rohit Girdhar, Rohit Patel, Romain Sauvestre, Ronnie Polidoro, Roshan Sumbaly, Ross Taylor, Ruan Silva, Rui Hou, Rui Wang, Saghar Hosseini, Sahana Chennabasappa, Sanjay Singh, Sean Bell, Seohyun~Sonia Kim, Sergey Edunov, Shaoliang Nie, Sharan Narang, Sharath Raparthy, Sheng Shen, Shengye Wan, Shruti Bhosale, Shun Zhang, Simon Vandenhende, Soumya Batra, Spencer Whitman, Sten Sootla, Stephane Collot, Suchin Gururangan, Sydney Borodinsky, Tamar Herman, Tara Fowler, Tarek Sheasha, Thomas Georgiou, Thomas Scialom, Tobias Speckbacher, Todor Mihaylov, Tong Xiao, Ujjwal Karn, Vedanuj Goswami, Vibhor Gupta, Vignesh Ramanathan, Viktor Kerkez, Vincent Gonguet, Virginie Do, Vish Vogeti, Vítor Albiero, Vladan Petrovic, Weiwei Chu, Wenhan Xiong, Wenyin Fu, Whitney Meers, Xavier Martinet, Xiaodong Wang, Xiaofang Wang, Xiaoqing~Ellen Tan, Xide Xia, Xinfeng Xie, Xuchao Jia, Xuewei Wang, Yaelle Goldschlag, Yashesh Gaur, Yasmine Babaei, Yi~Wen, Yiwen Song, Yuchen Zhang, Yue Li, Yuning Mao, Zacharie~Delpierre Coudert, Zheng Yan, Zhengxing Chen, Zoe Papakipos, Aaditya Singh, Aayushi Srivastava, Abha Jain, Adam Kelsey, Adam Shajnfeld, Adithya Gangidi, Adolfo Victoria, Ahuva Goldstand, Ajay Menon, Ajay Sharma, Alex Boesenberg, Alexei Baevski, Allie Feinstein, Amanda Kallet, Amit Sangani, Amos Teo, Anam Yunus, Andrei Lupu, Andres Alvarado, Andrew Caples, Andrew Gu, Andrew Ho, Andrew Poulton, Andrew Ryan, Ankit Ramchandani, Annie Dong, Annie Franco, Anuj Goyal, Aparajita Saraf, Arkabandhu Chowdhury, Ashley Gabriel, Ashwin Bharambe, Assaf Eisenman, Azadeh Yazdan, Beau James, Ben Maurer, Benjamin Leonhardi, Bernie Huang, Beth Loyd, Beto~De Paola, Bhargavi Paranjape, Bing Liu, Bo~Wu, Boyu Ni, Braden Hancock, Bram Wasti, Brandon Spence, Brani Stojkovic, Brian Gamido, Britt Montalvo, Carl Parker, Carly Burton, Catalina Mejia, Ce~Liu, Changhan Wang, Changkyu Kim, Chao Zhou, Chester Hu, Ching-Hsiang Chu, Chris Cai, Chris Tindal, Christoph Feichtenhofer, Cynthia Gao, Damon Civin, Dana Beaty, Daniel Kreymer, Daniel Li, David Adkins, David Xu, Davide Testuggine, Delia David, Devi Parikh, Diana Liskovich, Didem Foss, Dingkang Wang, Duc Le, Dustin Holland, Edward Dowling, Eissa Jamil, Elaine Montgomery, Eleonora Presani, Emily Hahn, Emily Wood, Eric-Tuan Le, Erik Brinkman, Esteban Arcaute, Evan Dunbar, Evan Smothers, Fei Sun, Felix Kreuk, Feng Tian, Filippos Kokkinos, Firat Ozgenel, Francesco Caggioni, Frank Kanayet, Frank Seide, Gabriela~Medina Florez, Gabriella Schwarz, Gada Badeer, Georgia Swee, Gil Halpern, Grant Herman, Grigory Sizov, Guangyi, Zhang, Guna Lakshminarayanan, Hakan Inan, Hamid Shojanazeri, Han Zou, Hannah Wang, Hanwen Zha, Haroun Habeeb, Harrison Rudolph, Helen Suk, Henry Aspegren, Hunter Goldman, Hongyuan Zhan, Ibrahim Damlaj, Igor Molybog, Igor Tufanov, Ilias Leontiadis, Irina-Elena Veliche, Itai Gat, Jake Weissman, James Geboski, James Kohli, Janice Lam, Japhet Asher, Jean-Baptiste Gaya, Jeff Marcus, Jeff Tang, Jennifer Chan, Jenny Zhen, Jeremy Reizenstein, Jeremy Teboul, Jessica Zhong, Jian Jin, Jingyi Yang, Joe Cummings, Jon Carvill, Jon Shepard, Jonathan McPhie, Jonathan Torres, Josh Ginsburg, Junjie Wang, Kai Wu, Kam~Hou U, Karan Saxena, Kartikay Khandelwal, Katayoun Zand, Kathy Matosich, Kaushik Veeraraghavan, Kelly Michelena, Keqian Li, Kiran Jagadeesh, Kun Huang, Kunal Chawla, Kyle Huang, Lailin Chen, Lakshya Garg, Lavender A, Leandro Silva, Lee Bell, Lei Zhang, Liangpeng Guo, Licheng Yu, Liron Moshkovich, Luca Wehrstedt, Madian Khabsa, Manav Avalani, Manish Bhatt, Martynas Mankus, Matan Hasson, Matthew Lennie, Matthias Reso, Maxim Groshev, Maxim Naumov, Maya Lathi, Meghan Keneally, Miao Liu, Michael~L. Seltzer, Michal Valko, Michelle Restrepo, Mihir Patel, Mik Vyatskov, Mikayel Samvelyan, Mike Clark, Mike Macey, Mike Wang, Miquel~Jubert Hermoso, Mo~Metanat, Mohammad Rastegari, Munish Bansal, Nandhini Santhanam, Natascha Parks, Natasha White, Navyata Bawa, Nayan Singhal, Nick Egebo, Nicolas Usunier, Nikhil Mehta, Nikolay~Pavlovich Laptev, Ning Dong, Norman Cheng, Oleg Chernoguz, Olivia Hart, Omkar Salpekar, Ozlem Kalinli, Parkin Kent, Parth Parekh, Paul Saab, Pavan Balaji, Pedro Rittner, Philip Bontrager, Pierre Roux, Piotr Dollar, Polina Zvyagina, Prashant Ratanchandani, Pritish Yuvraj, Qian Liang, Rachad Alao, Rachel Rodriguez, Rafi Ayub, Raghotham Murthy, Raghu Nayani, Rahul Mitra, Rangaprabhu Parthasarathy, Raymond Li, Rebekkah Hogan, Robin Battey, Rocky Wang, Russ Howes, Ruty Rinott, Sachin Mehta, Sachin Siby, Sai~Jayesh Bondu, Samyak Datta, Sara Chugh, Sara Hunt, Sargun Dhillon, Sasha Sidorov, Satadru Pan, Saurabh Mahajan, Saurabh Verma, Seiji Yamamoto, Sharadh Ramaswamy, Shaun Lindsay, Shaun Lindsay, Sheng Feng, Shenghao Lin, Shengxin~Cindy Zha, Shishir Patil, Shiva Shankar, Shuqiang Zhang, Shuqiang Zhang, Sinong Wang, Sneha Agarwal, Soji Sajuyigbe, Soumith Chintala, Stephanie Max, Stephen Chen, Steve Kehoe, Steve Satterfield, Sudarshan Govindaprasad, Sumit Gupta, Summer Deng, Sungmin Cho, Sunny Virk, Suraj Subramanian, Sy~Choudhury, Sydney Goldman, Tal Remez, Tamar Glaser, Tamara Best, Thilo Koehler, Thomas Robinson, Tianhe Li, Tianjun Zhang, Tim Matthews, Timothy Chou, Tzook Shaked, Varun Vontimitta, Victoria Ajayi, Victoria Montanez, Vijai Mohan, Vinay~Satish Kumar, Vishal Mangla, Vlad Ionescu, Vlad Poenaru, Vlad~Tiberiu Mihailescu, Vladimir Ivanov, Wei Li, Wenchen Wang, Wenwen Jiang, Wes Bouaziz, Will Constable, Xiaocheng Tang, Xiaojian Wu, Xiaolan Wang, Xilun Wu, Xinbo Gao, Yaniv Kleinman, Yanjun Chen, Ye~Hu, Ye~Jia, Ye~Qi, Yenda Li, Yilin Zhang, Ying Zhang, Yossi Adi, Youngjin Nam, Yu, Wang, Yu~Zhao, Yuchen Hao, Yundi Qian, Yunlu Li, Yuzi He, Zach Rait, Zachary DeVito, Zef Rosnbrick, Zhaoduo Wen, Zhenyu Yang, Zhiwei Zhao, and Zhiyu Ma.
\newblock The llama 3 herd of models, 2024.
\newblock URL \url{https://arxiv.org/abs/2407.21783}.

\bibitem[DeepSeek-AI et~al.(2025{\natexlab{a}})DeepSeek-AI, Liu, Feng, Xue, Wang, Wu, Lu, Zhao, Deng, Zhang, Ruan, Dai, Guo, Yang, Chen, Ji, Li, Lin, Dai, Luo, Hao, Chen, Li, Zhang, Bao, Xu, Wang, Zhang, Ding, Xin, Gao, Li, Qu, Cai, Liang, Guo, Ni, Li, Wang, Chen, Chen, Yuan, Qiu, Li, Song, Dong, Hu, Gao, Guan, Huang, Yu, Wang, Zhang, Xu, Xia, Zhao, Wang, Zhang, Li, Wang, Zhang, Zhang, Tang, Li, Tian, Huang, Wang, Zhang, Wang, Zhu, Chen, Du, Chen, Jin, Ge, Zhang, Pan, Wang, Xu, Zhang, Chen, Li, Lu, Zhou, Chen, Wu, Ye, Ye, Ma, Wang, Zhou, Yu, Zhou, Pan, Wang, Yun, Pei, Sun, Xiao, Zeng, Zhao, An, Liu, Liang, Gao, Yu, Zhang, Li, Jin, Wang, Bi, Liu, Wang, Shen, Chen, Zhang, Chen, Nie, Sun, Wang, Cheng, Liu, Xie, Liu, Yu, Song, Shan, Zhou, Yang, Li, Su, Lin, Li, Wang, Wei, Zhu, Zhang, Xu, Xu, Huang, Li, Zhao, Sun, Li, Wang, Yu, Zheng, Zhang, Shi, Xiong, He, Tang, Piao, Wang, Tan, Ma, Liu, Guo, Wu, Ou, Zhu, Wang, Gong, Zou, He, Zha, Xiong, Ma, Yan, Luo, You, Liu, Zhou, Wu, Ren, Ren, Sha, Fu, Xu, Huang, Zhang, Xie, Zhang, Hao, Gou, Ma, Yan, Shao, Xu, Wu, Zhang, Li, Gu, Zhu, Liu, Li, Xie, Song, Gao, and Pan]{dseekv3}
DeepSeek-AI, Aixin Liu, Bei Feng, Bing Xue, Bingxuan Wang, Bochao Wu, Chengda Lu, Chenggang Zhao, Chengqi Deng, Chenyu Zhang, Chong Ruan, Damai Dai, Daya Guo, Dejian Yang, Deli Chen, Dongjie Ji, Erhang Li, Fangyun Lin, Fucong Dai, Fuli Luo, Guangbo Hao, Guanting Chen, Guowei Li, H.~Zhang, Han Bao, Hanwei Xu, Haocheng Wang, Haowei Zhang, Honghui Ding, Huajian Xin, Huazuo Gao, Hui Li, Hui Qu, J.~L. Cai, Jian Liang, Jianzhong Guo, Jiaqi Ni, Jiashi Li, Jiawei Wang, Jin Chen, Jingchang Chen, Jingyang Yuan, Junjie Qiu, Junlong Li, Junxiao Song, Kai Dong, Kai Hu, Kaige Gao, Kang Guan, Kexin Huang, Kuai Yu, Lean Wang, Lecong Zhang, Lei Xu, Leyi Xia, Liang Zhao, Litong Wang, Liyue Zhang, Meng Li, Miaojun Wang, Mingchuan Zhang, Minghua Zhang, Minghui Tang, Mingming Li, Ning Tian, Panpan Huang, Peiyi Wang, Peng Zhang, Qiancheng Wang, Qihao Zhu, Qinyu Chen, Qiushi Du, R.~J. Chen, R.~L. Jin, Ruiqi Ge, Ruisong Zhang, Ruizhe Pan, Runji Wang, Runxin Xu, Ruoyu Zhang, Ruyi Chen, S.~S. Li, Shanghao Lu, Shangyan Zhou, Shanhuang Chen, Shaoqing Wu, Shengfeng Ye, Shengfeng Ye, Shirong Ma, Shiyu Wang, Shuang Zhou, Shuiping Yu, Shunfeng Zhou, Shuting Pan, T.~Wang, Tao Yun, Tian Pei, Tianyu Sun, W.~L. Xiao, Wangding Zeng, Wanjia Zhao, Wei An, Wen Liu, Wenfeng Liang, Wenjun Gao, Wenqin Yu, Wentao Zhang, X.~Q. Li, Xiangyue Jin, Xianzu Wang, Xiao Bi, Xiaodong Liu, Xiaohan Wang, Xiaojin Shen, Xiaokang Chen, Xiaokang Zhang, Xiaosha Chen, Xiaotao Nie, Xiaowen Sun, Xiaoxiang Wang, Xin Cheng, Xin Liu, Xin Xie, Xingchao Liu, Xingkai Yu, Xinnan Song, Xinxia Shan, Xinyi Zhou, Xinyu Yang, Xinyuan Li, Xuecheng Su, Xuheng Lin, Y.~K. Li, Y.~Q. Wang, Y.~X. Wei, Y.~X. Zhu, Yang Zhang, Yanhong Xu, Yanhong Xu, Yanping Huang, Yao Li, Yao Zhao, Yaofeng Sun, Yaohui Li, Yaohui Wang, Yi~Yu, Yi~Zheng, Yichao Zhang, Yifan Shi, Yiliang Xiong, Ying He, Ying Tang, Yishi Piao, Yisong Wang, Yixuan Tan, Yiyang Ma, Yiyuan Liu, Yongqiang Guo, Yu~Wu, Yuan Ou, Yuchen Zhu, Yuduan Wang, Yue Gong, Yuheng Zou, Yujia He, Yukun Zha, Yunfan Xiong, Yunxian Ma, Yuting Yan, Yuxiang Luo, Yuxiang You, Yuxuan Liu, Yuyang Zhou, Z.~F. Wu, Z.~Z. Ren, Zehui Ren, Zhangli Sha, Zhe Fu, Zhean Xu, Zhen Huang, Zhen Zhang, Zhenda Xie, Zhengyan Zhang, Zhewen Hao, Zhibin Gou, Zhicheng Ma, Zhigang Yan, Zhihong Shao, Zhipeng Xu, Zhiyu Wu, Zhongyu Zhang, Zhuoshu Li, Zihui Gu, Zijia Zhu, Zijun Liu, Zilin Li, Ziwei Xie, Ziyang Song, Ziyi Gao, and Zizheng Pan.
\newblock Deepseek-v3 technical report, 2025{\natexlab{a}}.
\newblock URL \url{https://arxiv.org/abs/2412.19437}.

\bibitem[Team et~al.(2025{\natexlab{a}})Team, Anil, Borgeaud, Alayrac, Yu, Soricut, Schalkwyk, Dai, Hauth, Millican, Silver, Johnson, Antonoglou, Schrittwieser, Glaese, Chen, Pitler, Lillicrap, Lazaridou, Firat, Molloy, Isard, Barham, Hennigan, Lee, Viola, Reynolds, Xu, Doherty, Collins, Meyer, Rutherford, Moreira, Ayoub, Goel, Krawczyk, Du, Chi, Cheng, Ni, Shah, Kane, Chan, Faruqui, Severyn, Lin, Li, Cheng, Ittycheriah, Mahdieh, Chen, Sun, Tran, Bagri, Lakshminarayanan, Liu, Orban, Güra, Zhou, Song, Boffy, Ganapathy, Zheng, Choe, Ágoston Weisz, Zhu, Lu, Gopal, Kahn, Kula, Pitman, Shah, Taropa, Merey, Baeuml, Chen, Shafey, Zhang, Sercinoglu, Tucker, Piqueras, Krikun, Barr, Savinov, Danihelka, Roelofs, White, Andreassen, von Glehn, Yagati, Kazemi, Gonzalez, Khalman, Sygnowski, Frechette, Smith, Culp, Proleev, Luan, Chen, Lottes, Schucher, Lebron, Rrustemi, Clay, Crone, Kocisky, Zhao, Perz, Yu, Howard, Bloniarz, Rae, Lu, Sifre, Maggioni, Alcober, Garrette, Barnes, Thakoor, Austin, Barth-Maron, Wong, Joshi, Chaabouni, Fatiha, Ahuja, Tomar, Senter, Chadwick, Kornakov, Attaluri, Iturrate, Liu, Li, Cogan, Chen, Jia, Gu, Zhang, Grimstad, Hartman, Garcia, Pillai, Devlin, Laskin, de~Las~Casas, Valter, Tao, Blanco, Badia, Reitter, Chen, Brennan, Rivera, Brin, Iqbal, Surita, Labanowski, Rao, Winkler, Parisotto, Gu, Olszewska, Addanki, Miech, Louis, Teplyashin, Brown, Catt, Balaguer, Xiang, Wang, Ashwood, Briukhov, Webson, Ganapathy, Sanghavi, Kannan, Chang, Stjerngren, Djolonga, Sun, Bapna, Aitchison, Pejman, Michalewski, Yu, Wang, Love, Ahn, Bloxwich, Han, Humphreys, Sellam, Bradbury, Godbole, Samangooei, Damoc, Kaskasoli, Arnold, Vasudevan, Agrawal, Riesa, Lepikhin, Tanburn, Srinivasan, Lim, Hodkinson, Shyam, Ferret, Hand, Garg, Paine, Li, Li, Giang, Neitz, Abbas, York, Reid, Cole, Chowdhery, Das, Rogozińska, Nikolaev, Sprechmann, Nado, Zilka, Prost, He, Monteiro, Mishra, Welty, Newlan, Jia, Allamanis, Hu, de~Liedekerke, Gilmer, Saroufim, Rijhwani, Hou, Shrivastava, Baddepudi, Goldin, Ozturel, Cassirer, Xu, Sohn, Sachan, Amplayo, Swanson, Petrova, Narayan, Guez, Brahma, Landon, Patel, Zhao, Villela, Wang, Jia, Rahtz, Giménez, Yeung, Keeling, Georgiev, Mincu, Wu, Haykal, Saputro, Vodrahalli, Qin, Cankara, Sharma, Fernando, Hawkins, Neyshabur, Kim, Hutter, Agrawal, Castro-Ros, van~den Driessche, Wang, Yang, yiin Chang, Komarek, McIlroy, Lučić, Zhang, Farhan, Sharman, Natsev, Michel, Bansal, Qiao, Cao, Shakeri, Butterfield, Chung, Rubenstein, Agrawal, Mensch, Soparkar, Lenc, Chung, Pope, Maggiore, Kay, Jhakra, Wang, Maynez, Phuong, Tobin, Tacchetti, Trebacz, Robinson, Katariya, Riedel, Bailey, Xiao, Ghelani, Aroyo, Slone, Houlsby, Xiong, Yang, Gribovskaya, Adler, Wirth, Lee, Li, Kagohara, Pavagadhi, Bridgers, Bortsova, Ghemawat, Ahmed, Liu, Powell, Bolina, Iinuma, Zablotskaia, Besley, Chung, Dozat, Comanescu, Si, Greer, Su, Polacek, Kaufman, Tokumine, Hu, Buchatskaya, Miao, Elhawaty, Siddhant, Tomasev, Xing, Greer, Miller, Ashraf, Roy, Zhang, Ma, Filos, Besta, Blevins, Klimenko, Yeh, Changpinyo, Mu, Chang, Pajarskas, Muir, Cohen, Lan, Haridasan, Marathe, Hansen, Douglas, Samuel, Wang, Austin, Lan, Jiang, Chiu, Lorenzo, Sjösund, Cevey, Gleicher, Avrahami, Boral, Srinivasan, Selo, May, Aisopos, Hussenot, Soares, Baumli, Chang, Recasens, Caine, Pritzel, Pavetic, Pardo, Gergely, Frye, Ramasesh, Horgan, Badola, Kassner, Roy, Dyer, Campos, Tomala, Tang, Badawy, White, Mustafa, Lang, Jindal, Vikram, Gong, Caelles, Hemsley, Thornton, Feng, Stokowiec, Zheng, Thacker, Çağlar Ünlü, Zhang, Saleh, Svensson, Bileschi, Patil, Anand, Ring, Tsihlas, Vezer, Selvi, Shevlane, Rodriguez, Kwiatkowski, Daruki, Rong, Dafoe, FitzGerald, Gu-Lemberg, Khan, Hendricks, Pellat, Feinberg, Cobon-Kerr, Sainath, Rauh, Hashemi, Ives, Hasson, Noland, Cao, Byrd, Hou, Wang, Sottiaux, Paganini, Lespiau, Moufarek, Hassan, Shivakumar, van Amersfoort, Mandhane, Joshi, Goyal, Tung, Brock, Sheahan, Misra, Li, Rakićević, Dehghani, Liu, Mittal, Oh, Noury, Sezener, Huot, Lamm, Cao, Chen, Mudgal, Stella, Brooks, Vasudevan, Liu, Chain, Melinkeri, Cohen, Wang, Seymore, Zubkov, Goel, Yue, Krishnakumaran, Albert, Hurley, Sano, Mohananey, Joughin, Filonov, Kępa, Eldawy, Lim, Rishi, Badiezadegan, Bos, Chang, Jain, Padmanabhan, Puttagunta, Krishna, Baker, Kalb, Bedapudi, Kurzrok, Lei, Yu, Litvin, Zhou, Wu, Sobell, Siciliano, Papir, Neale, Bragagnolo, Toor, Chen, Anklin, Wang, Feng, Gholami, Ling, Liu, Walter, Moghaddam, Kishore, Adamek, Mercado, Mallinson, Wandekar, Cagle, Ofek, Garrido, Lombriser, Mukha, Sun, Mohammad, Matak, Qian, Peswani, Janus, Yuan, Schelin, David, Garg, He, Duzhyi, Älgmyr, Lottaz, Li, Yadav, Xu, Chinien, Shivanna, Chuklin, Li, Spadine, Wolfe, Mohamed, Das, Dai, He, von Dincklage, Upadhyay, Maurya, Chi, Krause, Salama, Rabinovitch, M, Selvan, Dektiarev, Ghiasi, Guven, Gupta, Liu, Sharma, Shtacher, Paul, Akerlund, Aubet, Huang, Zhu, Zhu, Teixeira, Fritze, Bertolini, Marinescu, Bölle, Paulus, Gupta, Latkar, Chang, Sanders, Wilson, Wu, Tan, Thiet, Doshi, Lall, Mishra, Chen, Luong, Benjamin, Lee, Andrejczuk, Rabiej, Ranjan, Styrc, Yin, Simon, Harriott, Bansal, Robsky, Bacon, Greene, Mirylenka, Zhou, Sarvana, Goyal, Andermatt, Siegler, Horn, Israel, Pongetti, Chen, Selvatici, Silva, Wang, Tolins, Guu, Yogev, Cai, Agostini, Shah, Nguyen, Donnaile, Pereira, Friso, Stambler, Kurzrok, Kuang, Romanikhin, Geller, Yan, Jang, Lee, Fica, Malmi, Tan, Banica, Balle, Pham, Huang, Avram, Shi, Singh, Hidey, Ahuja, Saxena, Dooley, Potharaju, O'Neill, Gokulchandran, Foley, Zhao, Dusenberry, Liu, Mehta, Kotikalapudi, Safranek-Shrader, Goodman, Kessinger, Globen, Kolhar, Gorgolewski, Ibrahim, Song, Eichenbaum, Brovelli, Potluri, Lahoti, Baetu, Ghorbani, Chen, Crawford, Pal, Sridhar, Gurita, Mujika, Petrovski, Cedoz, Li, Chen, Santo, Goyal, Punjabi, Kappaganthu, Kwak, LV, Velury, Choudhury, Hall, Shah, Figueira, Thomas, Lu, Zhou, Kumar, Jurdi, Chikkerur, Ma, Yu, Kwak, Ähdel, Rajayogam, Choma, Liu, Barua, Ji, Park, Hellendoorn, Bailey, Bilal, Zhou, Khatir, Sutton, Rzadkowski, Macintosh, Vij, Shagin, Medina, Liang, Zhou, Shah, Bi, Dankovics, Banga, Lehmann, Bredesen, Lin, Hoffmann, Lai, Chung, Yang, Balani, Bražinskas, Sozanschi, Hayes, Alcalde, Makarov, Chen, Stella, Snijders, Mandl, Kärrman, Nowak, Wu, Dyck, Vaidyanathan, R, Mallet, Rudominer, Johnston, Mittal, Udathu, Christensen, Verma, Irving, Santucci, Elsayed, Davoodi, Georgiev, Tenney, Hua, Cideron, Leurent, Alnahlawi, Georgescu, Wei, Zheng, Scandinaro, Jiang, Snoek, Sundararajan, Wang, Ontiveros, Karo, Cole, Rajashekhar, Tumeh, Ben-David, Jain, Uesato, Datta, Bunyan, Wu, Zhang, Stanczyk, Zhang, Steiner, Naskar, Azzam, Johnson, Paszke, Chiu, Elias, Mohiuddin, Muhammad, Miao, Lee, Vieillard, Park, Zhang, Stanway, Garmon, Karmarkar, Dong, Lee, Kumar, Zhou, Evens, Isaac, Irving, Loper, Fink, Arkatkar, Chen, Shafran, Petrychenko, Chen, Jia, Levskaya, Zhu, Grabowski, Mao, Magni, Yao, Snaider, Casagrande, Palmer, Suganthan, Castaño, Giannoumis, Kim, Rybiński, Sreevatsa, Prendki, Soergel, Goedeckemeyer, Gierke, Jafari, Gaba, Wiesner, Wright, Wei, Vashisht, Kulizhskaya, Hoover, Le, Li, Iwuanyanwu, Liu, Ramirez, Khorlin, Cui, LIN, Wu, Aguilar, Pallo, Chakladar, Perng, Abellan, Zhang, Dasgupta, Kushman, Penchev, Repina, Wu, van~der Weide, Ponnapalli, Kaplan, Simsa, Li, Dousse, Yang, Piper, Ie, Pasumarthi, Lintz, Vijayakumar, Andor, Valenzuela, Lui, Paduraru, Peng, Lee, Zhang, Greene, Nguyen, Kurylowicz, Hardin, Dixon, Janzer, Choo, Feng, Zhang, Singhal, Du, McKinnon, Antropova, Bolukbasi, Keller, Reid, Finchelstein, Raad, Crocker, Hawkins, Dadashi, Gaffney, Franko, Bulanova, Leblond, Chung, Askham, Cobo, Xu, Fischer, Xu, Sorokin, Alberti, Lin, Evans, Dimitriev, Forbes, Banarse, Tung, Omernick, Bishop, Sterneck, Jain, Xia, Amid, Piccinno, Wang, Banzal, Mankowitz, Polozov, Krakovna, Brown, Bateni, Duan, Firoiu, Thotakuri, Natan, Geist, tan Girgin, Li, Ye, Roval, Tojo, Kwong, Lee-Thorp, Yew, Sinopalnikov, Ramos, Mellor, Sharma, Wu, Miller, Sonnerat, Vnukov, Greig, Beattie, Caveness, Bai, Eisenschlos, Korchemniy, Tsai, Jasarevic, Kong, Dao, Zheng, Liu, Yang, Zhu, Teh, Sanmiya, Gladchenko, Trdin, Toyama, Rosen, Tavakkol, Xue, Elkind, Woodman, Carpenter, Papamakarios, Kemp, Kafle, Grunina, Sinha, Talbert, Wu, Owusu-Afriyie, Du, Thornton, Pont-Tuset, Narayana, Li, Fatehi, Wieting, Ajmeri, Uria, Ko, Knight, Héliou, Niu, Gu, Pang, Li, Levine, Stolovich, Santamaria-Fernandez, Goenka, Yustalim, Strudel, Elqursh, Deck, Lee, Li, Levin, Hoffmann, Holtmann-Rice, Bachem, Arora, Koh, Yeganeh, Põder, Tariq, Sun, Ionita, Seyedhosseini, Tafti, Liu, Gulati, Liu, Ye, Chrzaszcz, Wang, Sethi, Li, Brown, Singh, Fan, Parisi, Stanton, Koverkathu, Choquette-Choo, Li, Lu, Ittycheriah, Shroff, Varadarajan, Bahargam, Willoughby, Gaddy, Desjardins, Cornero, Robenek, Mittal, Albrecht, Shenoy, Moiseev, Jacobsson, Ghaffarkhah, Rivière, Walton, Crepy, Parrish, Zhou, Farabet, Radebaugh, Srinivasan, van~der Salm, Fidjeland, Scellato, Latorre-Chimoto, Klimczak-Plucińska, Bridson, de~Cesare, Hudson, Mendolicchio, Walker, Morris, Mauger, Guseynov, Reid, Odoom, Loher, Cotruta, Yenugula, Grewe, Petrushkina, Duerig, Sanchez, Yadlowsky, Shen, Globerson, Webb, Dua, Li, Bhupatiraju, Hurt, Qureshi, Agarwal, Shani, Eyal, Khare, Belle, Wang, Tekur, Kale, Wei, Sang, Saeta, Liechty, Sun, Zhao, Lee, Nayak, Fritz, Vuyyuru, Aslanides, Vyas, Wicke, Ma, Eltyshev, Martin, Cate, Manyika, Amiri, Kim, Xiong, Kang, Luisier, Tripuraneni, Madras, Guo, Waters, Wang, Ainslie, Baldridge, Zhang, Pruthi, Bauer, Yang, Mansour, Gelman, Xu, Polovets, Liu, Cai, Chen, Sheng, Xue, Ozair, Angermueller, Li, Sinha, Wang, Wiesinger, Koukoumidis, Tian, Iyer, Gurumurthy, Goldenson, Shah, Blake, Yu, Urbanowicz, Palomaki, Fernando, Durden, Mehta, Momchev, Rahimtoroghi, Georgaki, Raul, Ruder, Redshaw, Lee, Zhou, Jalan, Li, Hechtman, Schuh, Nasr, Milan, Mikulik, Franco, Green, Nguyen, Kelley, Mahendru, Hu, Howland, Vargas, Hui, Bansal, Rao, Ghiya, Wang, Ye, Sarr, Preston, Elish, Li, Kaku,
  Gupta, Pasupat, Juan, Someswar, M., Chen, Amini, Fabrikant, Chu, Dong, Muthal, Buthpitiya, Jauhari, Hua, Khandelwal, Hitron, Ren, Rinaldi, Drath, Dabush, Jiang, Godhia, Sachs, Chen, Fan, Taitelbaum, Noga, Dai, Wang, Liang, Hamer, Ferng, Elkind, Atias, Lee, Listík, Carlen, van~de Kerkhof, Pikus, Zaher, Müller, Zykova, Stefanec, Gatsko, Hirnschall, Sethi, Xu, Ahuja, Tsai, Stefanoiu, Feng, Dhandhania, Katyal, Gupta, Parulekar, Pitta, Zhao, Bhatia, Bhavnani, Alhadlaq, Li, Danenberg, Tu, Pine, Filippova, Ghosh, Limonchik, Urala, Lanka, Clive, Sun, Li, Wu, Hongtongsak, Li, Thakkar, Omarov, Majmundar, Alverson, Kucharski, Patel, Jain, Zabelin, Pelagatti, Kohli, Kumar, Kim, Sankar, Shah, Ramachandruni, Zeng, Bariach, Weidinger, Vu, Andreev, He, Hui, Kashem, Subramanya, Hsiao, Hassabis, Kavukcuoglu, Sadovsky, Le, Strohman, Wu, Petrov, Dean, and Vinyals]{googlegemini}
Gemini Team, Rohan Anil, Sebastian Borgeaud, Jean-Baptiste Alayrac, Jiahui Yu, Radu Soricut, Johan Schalkwyk, Andrew~M. Dai, Anja Hauth, Katie Millican, David Silver, Melvin Johnson, Ioannis Antonoglou, Julian Schrittwieser, Amelia Glaese, Jilin Chen, Emily Pitler, Timothy Lillicrap, Angeliki Lazaridou, Orhan Firat, James Molloy, Michael Isard, Paul~R. Barham, Tom Hennigan, Benjamin Lee, Fabio Viola, Malcolm Reynolds, Yuanzhong Xu, Ryan Doherty, Eli Collins, Clemens Meyer, Eliza Rutherford, Erica Moreira, Kareem Ayoub, Megha Goel, Jack Krawczyk, Cosmo Du, Ed~Chi, Heng-Tze Cheng, Eric Ni, Purvi Shah, Patrick Kane, Betty Chan, Manaal Faruqui, Aliaksei Severyn, Hanzhao Lin, YaGuang Li, Yong Cheng, Abe Ittycheriah, Mahdis Mahdieh, Mia Chen, Pei Sun, Dustin Tran, Sumit Bagri, Balaji Lakshminarayanan, Jeremiah Liu, Andras Orban, Fabian Güra, Hao Zhou, Xinying Song, Aurelien Boffy, Harish Ganapathy, Steven Zheng, HyunJeong Choe, Ágoston Weisz, Tao Zhu, Yifeng Lu, Siddharth Gopal, Jarrod Kahn, Maciej Kula, Jeff Pitman, Rushin Shah, Emanuel Taropa, Majd~Al Merey, Martin Baeuml, Zhifeng Chen, Laurent~El Shafey, Yujing Zhang, Olcan Sercinoglu, George Tucker, Enrique Piqueras, Maxim Krikun, Iain Barr, Nikolay Savinov, Ivo Danihelka, Becca Roelofs, Anaïs White, Anders Andreassen, Tamara von Glehn, Lakshman Yagati, Mehran Kazemi, Lucas Gonzalez, Misha Khalman, Jakub Sygnowski, Alexandre Frechette, Charlotte Smith, Laura Culp, Lev Proleev, Yi~Luan, Xi~Chen, James Lottes, Nathan Schucher, Federico Lebron, Alban Rrustemi, Natalie Clay, Phil Crone, Tomas Kocisky, Jeffrey Zhao, Bartek Perz, Dian Yu, Heidi Howard, Adam Bloniarz, Jack~W. Rae, Han Lu, Laurent Sifre, Marcello Maggioni, Fred Alcober, Dan Garrette, Megan Barnes, Shantanu Thakoor, Jacob Austin, Gabriel Barth-Maron, William Wong, Rishabh Joshi, Rahma Chaabouni, Deeni Fatiha, Arun Ahuja, Gaurav~Singh Tomar, Evan Senter, Martin Chadwick, Ilya Kornakov, Nithya Attaluri, Iñaki Iturrate, Ruibo Liu, Yunxuan Li, Sarah Cogan, Jeremy Chen, Chao Jia, Chenjie Gu, Qiao Zhang, Jordan Grimstad, Ale~Jakse Hartman, Xavier Garcia, Thanumalayan~Sankaranarayana Pillai, Jacob Devlin, Michael Laskin, Diego de~Las~Casas, Dasha Valter, Connie Tao, Lorenzo Blanco, Adrià~Puigdomènech Badia, David Reitter, Mianna Chen, Jenny Brennan, Clara Rivera, Sergey Brin, Shariq Iqbal, Gabriela Surita, Jane Labanowski, Abhi Rao, Stephanie Winkler, Emilio Parisotto, Yiming Gu, Kate Olszewska, Ravi Addanki, Antoine Miech, Annie Louis, Denis Teplyashin, Geoff Brown, Elliot Catt, Jan Balaguer, Jackie Xiang, Pidong Wang, Zoe Ashwood, Anton Briukhov, Albert Webson, Sanjay Ganapathy, Smit Sanghavi, Ajay Kannan, Ming-Wei Chang, Axel Stjerngren, Josip Djolonga, Yuting Sun, Ankur Bapna, Matthew Aitchison, Pedram Pejman, Henryk Michalewski, Tianhe Yu, Cindy Wang, Juliette Love, Junwhan Ahn, Dawn Bloxwich, Kehang Han, Peter Humphreys, Thibault Sellam, James Bradbury, Varun Godbole, Sina Samangooei, Bogdan Damoc, Alex Kaskasoli, Sébastien M.~R. Arnold, Vijay Vasudevan, Shubham Agrawal, Jason Riesa, Dmitry Lepikhin, Richard Tanburn, Srivatsan Srinivasan, Hyeontaek Lim, Sarah Hodkinson, Pranav Shyam, Johan Ferret, Steven Hand, Ankush Garg, Tom~Le Paine, Jian Li, Yujia Li, Minh Giang, Alexander Neitz, Zaheer Abbas, Sarah York, Machel Reid, Elizabeth Cole, Aakanksha Chowdhery, Dipanjan Das, Dominika Rogozińska, Vitaliy Nikolaev, Pablo Sprechmann, Zachary Nado, Lukas Zilka, Flavien Prost, Luheng He, Marianne Monteiro, Gaurav Mishra, Chris Welty, Josh Newlan, Dawei Jia, Miltiadis Allamanis, Clara~Huiyi Hu, Raoul de~Liedekerke, Justin Gilmer, Carl Saroufim, Shruti Rijhwani, Shaobo Hou, Disha Shrivastava, Anirudh Baddepudi, Alex Goldin, Adnan Ozturel, Albin Cassirer, Yunhan Xu, Daniel Sohn, Devendra Sachan, Reinald~Kim Amplayo, Craig Swanson, Dessie Petrova, Shashi Narayan, Arthur Guez, Siddhartha Brahma, Jessica Landon, Miteyan Patel, Ruizhe Zhao, Kevin Villela, Luyu Wang, Wenhao Jia, Matthew Rahtz, Mai Giménez, Legg Yeung, James Keeling, Petko Georgiev, Diana Mincu, Boxi Wu, Salem Haykal, Rachel Saputro, Kiran Vodrahalli, James Qin, Zeynep Cankara, Abhanshu Sharma, Nick Fernando, Will Hawkins, Behnam Neyshabur, Solomon Kim, Adrian Hutter, Priyanka Agrawal, Alex Castro-Ros, George van~den Driessche, Tao Wang, Fan Yang, Shuo yiin Chang, Paul Komarek, Ross McIlroy, Mario Lučić, Guodong Zhang, Wael Farhan, Michael Sharman, Paul Natsev, Paul Michel, Yamini Bansal, Siyuan Qiao, Kris Cao, Siamak Shakeri, Christina Butterfield, Justin Chung, Paul~Kishan Rubenstein, Shivani Agrawal, Arthur Mensch, Kedar Soparkar, Karel Lenc, Timothy Chung, Aedan Pope, Loren Maggiore, Jackie Kay, Priya Jhakra, Shibo Wang, Joshua Maynez, Mary Phuong, Taylor Tobin, Andrea Tacchetti, Maja Trebacz, Kevin Robinson, Yash Katariya, Sebastian Riedel, Paige Bailey, Kefan Xiao, Nimesh Ghelani, Lora Aroyo, Ambrose Slone, Neil Houlsby, Xuehan Xiong, Zhen Yang, Elena Gribovskaya, Jonas Adler, Mateo Wirth, Lisa Lee, Music Li, Thais Kagohara, Jay Pavagadhi, Sophie Bridgers, Anna Bortsova, Sanjay Ghemawat, Zafarali Ahmed, Tianqi Liu, Richard Powell, Vijay Bolina, Mariko Iinuma, Polina Zablotskaia, James Besley, Da-Woon Chung, Timothy Dozat, Ramona Comanescu, Xiance Si, Jeremy Greer, Guolong Su, Martin Polacek, Raphaël~Lopez Kaufman, Simon Tokumine, Hexiang Hu, Elena Buchatskaya, Yingjie Miao, Mohamed Elhawaty, Aditya Siddhant, Nenad Tomasev, Jinwei Xing, Christina Greer, Helen Miller, Shereen Ashraf, Aurko Roy, Zizhao Zhang, Ada Ma, Angelos Filos, Milos Besta, Rory Blevins, Ted Klimenko, Chih-Kuan Yeh, Soravit Changpinyo, Jiaqi Mu, Oscar Chang, Mantas Pajarskas, Carrie Muir, Vered Cohen, Charline~Le Lan, Krishna Haridasan, Amit Marathe, Steven Hansen, Sholto Douglas, Rajkumar Samuel, Mingqiu Wang, Sophia Austin, Chang Lan, Jiepu Jiang, Justin Chiu, Jaime~Alonso Lorenzo, Lars~Lowe Sjösund, Sébastien Cevey, Zach Gleicher, Thi Avrahami, Anudhyan Boral, Hansa Srinivasan, Vittorio Selo, Rhys May, Konstantinos Aisopos, Léonard Hussenot, Livio~Baldini Soares, Kate Baumli, Michael~B. Chang, Adrià Recasens, Ben Caine, Alexander Pritzel, Filip Pavetic, Fabio Pardo, Anita Gergely, Justin Frye, Vinay Ramasesh, Dan Horgan, Kartikeya Badola, Nora Kassner, Subhrajit Roy, Ethan Dyer, Víctor~Campos Campos, Alex Tomala, Yunhao Tang, Dalia~El Badawy, Elspeth White, Basil Mustafa, Oran Lang, Abhishek Jindal, Sharad Vikram, Zhitao Gong, Sergi Caelles, Ross Hemsley, Gregory Thornton, Fangxiaoyu Feng, Wojciech Stokowiec, Ce~Zheng, Phoebe Thacker, Çağlar Ünlü, Zhishuai Zhang, Mohammad Saleh, James Svensson, Max Bileschi, Piyush Patil, Ankesh Anand, Roman Ring, Katerina Tsihlas, Arpi Vezer, Marco Selvi, Toby Shevlane, Mikel Rodriguez, Tom Kwiatkowski, Samira Daruki, Keran Rong, Allan Dafoe, Nicholas FitzGerald, Keren Gu-Lemberg, Mina Khan, Lisa~Anne Hendricks, Marie Pellat, Vladimir Feinberg, James Cobon-Kerr, Tara Sainath, Maribeth Rauh, Sayed~Hadi Hashemi, Richard Ives, Yana Hasson, Eric Noland, Yuan Cao, Nathan Byrd, Le~Hou, Qingze Wang, Thibault Sottiaux, Michela Paganini, Jean-Baptiste Lespiau, Alexandre Moufarek, Samer Hassan, Kaushik Shivakumar, Joost van Amersfoort, Amol Mandhane, Pratik Joshi, Anirudh Goyal, Matthew Tung, Andrew Brock, Hannah Sheahan, Vedant Misra, Cheng Li, Nemanja Rakićević, Mostafa Dehghani, Fangyu Liu, Sid Mittal, Junhyuk Oh, Seb Noury, Eren Sezener, Fantine Huot, Matthew Lamm, Nicola~De Cao, Charlie Chen, Sidharth Mudgal, Romina Stella, Kevin Brooks, Gautam Vasudevan, Chenxi Liu, Mainak Chain, Nivedita Melinkeri, Aaron Cohen, Venus Wang, Kristie Seymore, Sergey Zubkov, Rahul Goel, Summer Yue, Sai Krishnakumaran, Brian Albert, Nate Hurley, Motoki Sano, Anhad Mohananey, Jonah Joughin, Egor Filonov, Tomasz Kępa, Yomna Eldawy, Jiawern Lim, Rahul Rishi, Shirin Badiezadegan, Taylor Bos, Jerry Chang, Sanil Jain, Sri Gayatri~Sundara Padmanabhan, Subha Puttagunta, Kalpesh Krishna, Leslie Baker, Norbert Kalb, Vamsi Bedapudi, Adam Kurzrok, Shuntong Lei, Anthony Yu, Oren Litvin, Xiang Zhou, Zhichun Wu, Sam Sobell, Andrea Siciliano, Alan Papir, Robby Neale, Jonas Bragagnolo, Tej Toor, Tina Chen, Valentin Anklin, Feiran Wang, Richie Feng, Milad Gholami, Kevin Ling, Lijuan Liu, Jules Walter, Hamid Moghaddam, Arun Kishore, Jakub Adamek, Tyler Mercado, Jonathan Mallinson, Siddhinita Wandekar, Stephen Cagle, Eran Ofek, Guillermo Garrido, Clemens Lombriser, Maksim Mukha, Botu Sun, Hafeezul~Rahman Mohammad, Josip Matak, Yadi Qian, Vikas Peswani, Pawel Janus, Quan Yuan, Leif Schelin, Oana David, Ankur Garg, Yifan He, Oleksii Duzhyi, Anton Älgmyr, Timothée Lottaz, Qi~Li, Vikas Yadav, Luyao Xu, Alex Chinien, Rakesh Shivanna, Aleksandr Chuklin, Josie Li, Carrie Spadine, Travis Wolfe, Kareem Mohamed, Subhabrata Das, Zihang Dai, Kyle He, Daniel von Dincklage, Shyam Upadhyay, Akanksha Maurya, Luyan Chi, Sebastian Krause, Khalid Salama, Pam~G Rabinovitch, Pavan Kumar~Reddy M, Aarush Selvan, Mikhail Dektiarev, Golnaz Ghiasi, Erdem Guven, Himanshu Gupta, Boyi Liu, Deepak Sharma, Idan~Heimlich Shtacher, Shachi Paul, Oscar Akerlund, François-Xavier Aubet, Terry Huang, Chen Zhu, Eric Zhu, Elico Teixeira, Matthew Fritze, Francesco Bertolini, Liana-Eleonora Marinescu, Martin Bölle, Dominik Paulus, Khyatti Gupta, Tejasi Latkar, Max Chang, Jason Sanders, Roopa Wilson, Xuewei Wu, Yi-Xuan Tan, Lam~Nguyen Thiet, Tulsee Doshi, Sid Lall, Swaroop Mishra, Wanming Chen, Thang Luong, Seth Benjamin, Jasmine Lee, Ewa Andrejczuk, Dominik Rabiej, Vipul Ranjan, Krzysztof Styrc, Pengcheng Yin, Jon Simon, Malcolm~Rose Harriott, Mudit Bansal, Alexei Robsky, Geoff Bacon, David Greene, Daniil Mirylenka, Chen Zhou, Obaid Sarvana, Abhimanyu Goyal, Samuel Andermatt, Patrick Siegler, Ben Horn, Assaf Israel, Francesco Pongetti, Chih-Wei~"Louis" Chen, Marco Selvatici, Pedro Silva, Kathie Wang, Jackson Tolins, Kelvin Guu, Roey Yogev, Xiaochen Cai, Alessandro Agostini, Maulik Shah, Hung Nguyen, Noah~Ó Donnaile, Sébastien Pereira, Linda Friso, Adam Stambler, Adam Kurzrok, Chenkai Kuang, Yan
  Romanikhin, Mark Geller, ZJ~Yan, Kane Jang, Cheng-Chun Lee, Wojciech Fica, Eric Malmi, Qijun Tan, Dan Banica, Daniel Balle, Ryan Pham, Yanping Huang, Diana Avram, Hongzhi Shi, Jasjot Singh, Chris Hidey, Niharika Ahuja, Pranab Saxena, Dan Dooley, Srividya~Pranavi Potharaju, Eileen O'Neill, Anand Gokulchandran, Ryan Foley, Kai Zhao, Mike Dusenberry, Yuan Liu, Pulkit Mehta, Ragha Kotikalapudi, Chalence Safranek-Shrader, Andrew Goodman, Joshua Kessinger, Eran Globen, Prateek Kolhar, Chris Gorgolewski, Ali Ibrahim, Yang Song, Ali Eichenbaum, Thomas Brovelli, Sahitya Potluri, Preethi Lahoti, Cip Baetu, Ali Ghorbani, Charles Chen, Andy Crawford, Shalini Pal, Mukund Sridhar, Petru Gurita, Asier Mujika, Igor Petrovski, Pierre-Louis Cedoz, Chenmei Li, Shiyuan Chen, Niccolò~Dal Santo, Siddharth Goyal, Jitesh Punjabi, Karthik Kappaganthu, Chester Kwak, Pallavi LV, Sarmishta Velury, Himadri Choudhury, Jamie Hall, Premal Shah, Ricardo Figueira, Matt Thomas, Minjie Lu, Ting Zhou, Chintu Kumar, Thomas Jurdi, Sharat Chikkerur, Yenai Ma, Adams Yu, Soo Kwak, Victor Ähdel, Sujeevan Rajayogam, Travis Choma, Fei Liu, Aditya Barua, Colin Ji, Ji~Ho Park, Vincent Hellendoorn, Alex Bailey, Taylan Bilal, Huanjie Zhou, Mehrdad Khatir, Charles Sutton, Wojciech Rzadkowski, Fiona Macintosh, Roopali Vij, Konstantin Shagin, Paul Medina, Chen Liang, Jinjing Zhou, Pararth Shah, Yingying Bi, Attila Dankovics, Shipra Banga, Sabine Lehmann, Marissa Bredesen, Zifan Lin, John~Eric Hoffmann, Jonathan Lai, Raynald Chung, Kai Yang, Nihal Balani, Arthur Bražinskas, Andrei Sozanschi, Matthew Hayes, Héctor~Fernández Alcalde, Peter Makarov, Will Chen, Antonio Stella, Liselotte Snijders, Michael Mandl, Ante Kärrman, Paweł Nowak, Xinyi Wu, Alex Dyck, Krishnan Vaidyanathan, Raghavender R, Jessica Mallet, Mitch Rudominer, Eric Johnston, Sushil Mittal, Akhil Udathu, Janara Christensen, Vishal Verma, Zach Irving, Andreas Santucci, Gamaleldin Elsayed, Elnaz Davoodi, Marin Georgiev, Ian Tenney, Nan Hua, Geoffrey Cideron, Edouard Leurent, Mahmoud Alnahlawi, Ionut Georgescu, Nan Wei, Ivy Zheng, Dylan Scandinaro, Heinrich Jiang, Jasper Snoek, Mukund Sundararajan, Xuezhi Wang, Zack Ontiveros, Itay Karo, Jeremy Cole, Vinu Rajashekhar, Lara Tumeh, Eyal Ben-David, Rishub Jain, Jonathan Uesato, Romina Datta, Oskar Bunyan, Shimu Wu, John Zhang, Piotr Stanczyk, Ye~Zhang, David Steiner, Subhajit Naskar, Michael Azzam, Matthew Johnson, Adam Paszke, Chung-Cheng Chiu, Jaume~Sanchez Elias, Afroz Mohiuddin, Faizan Muhammad, Jin Miao, Andrew Lee, Nino Vieillard, Jane Park, Jiageng Zhang, Jeff Stanway, Drew Garmon, Abhijit Karmarkar, Zhe Dong, Jong Lee, Aviral Kumar, Luowei Zhou, Jonathan Evens, William Isaac, Geoffrey Irving, Edward Loper, Michael Fink, Isha Arkatkar, Nanxin Chen, Izhak Shafran, Ivan Petrychenko, Zhe Chen, Johnson Jia, Anselm Levskaya, Zhenkai Zhu, Peter Grabowski, Yu~Mao, Alberto Magni, Kaisheng Yao, Javier Snaider, Norman Casagrande, Evan Palmer, Paul Suganthan, Alfonso Castaño, Irene Giannoumis, Wooyeol Kim, Mikołaj Rybiński, Ashwin Sreevatsa, Jennifer Prendki, David Soergel, Adrian Goedeckemeyer, Willi Gierke, Mohsen Jafari, Meenu Gaba, Jeremy Wiesner, Diana~Gage Wright, Yawen Wei, Harsha Vashisht, Yana Kulizhskaya, Jay Hoover, Maigo Le, Lu~Li, Chimezie Iwuanyanwu, Lu~Liu, Kevin Ramirez, Andrey Khorlin, Albert Cui, Tian LIN, Marcus Wu, Ricardo Aguilar, Keith Pallo, Abhishek Chakladar, Ginger Perng, Elena~Allica Abellan, Mingyang Zhang, Ishita Dasgupta, Nate Kushman, Ivo Penchev, Alena Repina, Xihui Wu, Tom van~der Weide, Priya Ponnapalli, Caroline Kaplan, Jiri Simsa, Shuangfeng Li, Olivier Dousse, Fan Yang, Jeff Piper, Nathan Ie, Rama Pasumarthi, Nathan Lintz, Anitha Vijayakumar, Daniel Andor, Pedro Valenzuela, Minnie Lui, Cosmin Paduraru, Daiyi Peng, Katherine Lee, Shuyuan Zhang, Somer Greene, Duc~Dung Nguyen, Paula Kurylowicz, Cassidy Hardin, Lucas Dixon, Lili Janzer, Kiam Choo, Ziqiang Feng, Biao Zhang, Achintya Singhal, Dayou Du, Dan McKinnon, Natasha Antropova, Tolga Bolukbasi, Orgad Keller, David Reid, Daniel Finchelstein, Maria~Abi Raad, Remi Crocker, Peter Hawkins, Robert Dadashi, Colin Gaffney, Ken Franko, Anna Bulanova, Rémi Leblond, Shirley Chung, Harry Askham, Luis~C. Cobo, Kelvin Xu, Felix Fischer, Jun Xu, Christina Sorokin, Chris Alberti, Chu-Cheng Lin, Colin Evans, Alek Dimitriev, Hannah Forbes, Dylan Banarse, Zora Tung, Mark Omernick, Colton Bishop, Rachel Sterneck, Rohan Jain, Jiawei Xia, Ehsan Amid, Francesco Piccinno, Xingyu Wang, Praseem Banzal, Daniel~J. Mankowitz, Alex Polozov, Victoria Krakovna, Sasha Brown, MohammadHossein Bateni, Dennis Duan, Vlad Firoiu, Meghana Thotakuri, Tom Natan, Matthieu Geist, Ser tan Girgin, Hui Li, Jiayu Ye, Ofir Roval, Reiko Tojo, Michael Kwong, James Lee-Thorp, Christopher Yew, Danila Sinopalnikov, Sabela Ramos, John Mellor, Abhishek Sharma, Kathy Wu, David Miller, Nicolas Sonnerat, Denis Vnukov, Rory Greig, Jennifer Beattie, Emily Caveness, Libin Bai, Julian Eisenschlos, Alex Korchemniy, Tomy Tsai, Mimi Jasarevic, Weize Kong, Phuong Dao, Zeyu Zheng, Frederick Liu, Fan Yang, Rui Zhu, Tian~Huey Teh, Jason Sanmiya, Evgeny Gladchenko, Nejc Trdin, Daniel Toyama, Evan Rosen, Sasan Tavakkol, Linting Xue, Chen Elkind, Oliver Woodman, John Carpenter, George Papamakarios, Rupert Kemp, Sushant Kafle, Tanya Grunina, Rishika Sinha, Alice Talbert, Diane Wu, Denese Owusu-Afriyie, Cosmo Du, Chloe Thornton, Jordi Pont-Tuset, Pradyumna Narayana, Jing Li, Saaber Fatehi, John Wieting, Omar Ajmeri, Benigno Uria, Yeongil Ko, Laura Knight, Amélie Héliou, Ning Niu, Shane Gu, Chenxi Pang, Yeqing Li, Nir Levine, Ariel Stolovich, Rebeca Santamaria-Fernandez, Sonam Goenka, Wenny Yustalim, Robin Strudel, Ali Elqursh, Charlie Deck, Hyo Lee, Zonglin Li, Kyle Levin, Raphael Hoffmann, Dan Holtmann-Rice, Olivier Bachem, Sho Arora, Christy Koh, Soheil~Hassas Yeganeh, Siim Põder, Mukarram Tariq, Yanhua Sun, Lucian Ionita, Mojtaba Seyedhosseini, Pouya Tafti, Zhiyu Liu, Anmol Gulati, Jasmine Liu, Xinyu Ye, Bart Chrzaszcz, Lily Wang, Nikhil Sethi, Tianrun Li, Ben Brown, Shreya Singh, Wei Fan, Aaron Parisi, Joe Stanton, Vinod Koverkathu, Christopher~A. Choquette-Choo, Yunjie Li, TJ~Lu, Abe Ittycheriah, Prakash Shroff, Mani Varadarajan, Sanaz Bahargam, Rob Willoughby, David Gaddy, Guillaume Desjardins, Marco Cornero, Brona Robenek, Bhavishya Mittal, Ben Albrecht, Ashish Shenoy, Fedor Moiseev, Henrik Jacobsson, Alireza Ghaffarkhah, Morgane Rivière, Alanna Walton, Clément Crepy, Alicia Parrish, Zongwei Zhou, Clement Farabet, Carey Radebaugh, Praveen Srinivasan, Claudia van~der Salm, Andreas Fidjeland, Salvatore Scellato, Eri Latorre-Chimoto, Hanna Klimczak-Plucińska, David Bridson, Dario de~Cesare, Tom Hudson, Piermaria Mendolicchio, Lexi Walker, Alex Morris, Matthew Mauger, Alexey Guseynov, Alison Reid, Seth Odoom, Lucia Loher, Victor Cotruta, Madhavi Yenugula, Dominik Grewe, Anastasia Petrushkina, Tom Duerig, Antonio Sanchez, Steve Yadlowsky, Amy Shen, Amir Globerson, Lynette Webb, Sahil Dua, Dong Li, Surya Bhupatiraju, Dan Hurt, Haroon Qureshi, Ananth Agarwal, Tomer Shani, Matan Eyal, Anuj Khare, Shreyas~Rammohan Belle, Lei Wang, Chetan Tekur, Mihir~Sanjay Kale, Jinliang Wei, Ruoxin Sang, Brennan Saeta, Tyler Liechty, Yi~Sun, Yao Zhao, Stephan Lee, Pandu Nayak, Doug Fritz, Manish~Reddy Vuyyuru, John Aslanides, Nidhi Vyas, Martin Wicke, Xiao Ma, Evgenii Eltyshev, Nina Martin, Hardie Cate, James Manyika, Keyvan Amiri, Yelin Kim, Xi~Xiong, Kai Kang, Florian Luisier, Nilesh Tripuraneni, David Madras, Mandy Guo, Austin Waters, Oliver Wang, Joshua Ainslie, Jason Baldridge, Han Zhang, Garima Pruthi, Jakob Bauer, Feng Yang, Riham Mansour, Jason Gelman, Yang Xu, George Polovets, Ji~Liu, Honglong Cai, Warren Chen, XiangHai Sheng, Emily Xue, Sherjil Ozair, Christof Angermueller, Xiaowei Li, Anoop Sinha, Weiren Wang, Julia Wiesinger, Emmanouil Koukoumidis, Yuan Tian, Anand Iyer, Madhu Gurumurthy, Mark Goldenson, Parashar Shah, MK~Blake, Hongkun Yu, Anthony Urbanowicz, Jennimaria Palomaki, Chrisantha Fernando, Ken Durden, Harsh Mehta, Nikola Momchev, Elahe Rahimtoroghi, Maria Georgaki, Amit Raul, Sebastian Ruder, Morgan Redshaw, Jinhyuk Lee, Denny Zhou, Komal Jalan, Dinghua Li, Blake Hechtman, Parker Schuh, Milad Nasr, Kieran Milan, Vladimir Mikulik, Juliana Franco, Tim Green, Nam Nguyen, Joe Kelley, Aroma Mahendru, Andrea Hu, Joshua Howland, Ben Vargas, Jeffrey Hui, Kshitij Bansal, Vikram Rao, Rakesh Ghiya, Emma Wang, Ke~Ye, Jean~Michel Sarr, Melanie~Moranski Preston, Madeleine Elish, Steve Li, Aakash Kaku, Jigar Gupta, Ice Pasupat, Da-Cheng Juan, Milan Someswar, Tejvi M., Xinyun Chen, Aida Amini, Alex Fabrikant, Eric Chu, Xuanyi Dong, Amruta Muthal, Senaka Buthpitiya, Sarthak Jauhari, Nan Hua, Urvashi Khandelwal, Ayal Hitron, Jie Ren, Larissa Rinaldi, Shahar Drath, Avigail Dabush, Nan-Jiang Jiang, Harshal Godhia, Uli Sachs, Anthony Chen, Yicheng Fan, Hagai Taitelbaum, Hila Noga, Zhuyun Dai, James Wang, Chen Liang, Jenny Hamer, Chun-Sung Ferng, Chenel Elkind, Aviel Atias, Paulina Lee, Vít Listík, Mathias Carlen, Jan van~de Kerkhof, Marcin Pikus, Krunoslav Zaher, Paul Müller, Sasha Zykova, Richard Stefanec, Vitaly Gatsko, Christoph Hirnschall, Ashwin Sethi, Xingyu~Federico Xu, Chetan Ahuja, Beth Tsai, Anca Stefanoiu, Bo~Feng, Keshav Dhandhania, Manish Katyal, Akshay Gupta, Atharva Parulekar, Divya Pitta, Jing Zhao, Vivaan Bhatia, Yashodha Bhavnani, Omar Alhadlaq, Xiaolin Li, Peter Danenberg, Dennis Tu, Alex Pine, Vera Filippova, Abhipso Ghosh, Ben Limonchik, Bhargava Urala, Chaitanya~Krishna Lanka, Derik Clive, Yi~Sun, Edward Li, Hao Wu, Kevin Hongtongsak, Ianna Li, Kalind Thakkar, Kuanysh Omarov, Kushal Majmundar, Michael Alverson, Michael Kucharski, Mohak Patel, Mudit Jain, Maksim Zabelin, Paolo Pelagatti, Rohan Kohli, Saurabh Kumar, Joseph Kim, Swetha Sankar, Vineet Shah, Lakshmi Ramachandruni, Xiangkai Zeng, Ben Bariach, Laura Weidinger, Tu~Vu, Alek Andreev, Antoine He, Kevin Hui, Sheleem Kashem, Amar Subramanya, Sissie Hsiao, Demis
  Hassabis, Koray Kavukcuoglu, Adam Sadovsky, Quoc Le, Trevor Strohman, Yonghui Wu, Slav Petrov, Jeffrey Dean, and Oriol Vinyals.
\newblock Gemini: A family of highly capable multimodal models, 2025{\natexlab{a}}.
\newblock URL \url{https://arxiv.org/abs/2312.11805}.

\bibitem[Labs et~al.(2025)Labs, Batifol, Blattmann, Boesel, Consul, Diagne, Dockhorn, English, English, Esser, Kulal, Lacey, Levi, Li, Lorenz, Müller, Podell, Rombach, Saini, Sauer, and Smith]{flux}
Black~Forest Labs, Stephen Batifol, Andreas Blattmann, Frederic Boesel, Saksham Consul, Cyril Diagne, Tim Dockhorn, Jack English, Zion English, Patrick Esser, Sumith Kulal, Kyle Lacey, Yam Levi, Cheng Li, Dominik Lorenz, Jonas Müller, Dustin Podell, Robin Rombach, Harry Saini, Axel Sauer, and Luke Smith.
\newblock Flux.1 kontext: Flow matching for in-context image generation and editing in latent space, 2025.
\newblock URL \url{https://arxiv.org/abs/2506.15742}.

\bibitem[Ye et~al.(2023)Ye, Chen, Xu, Zu, Shao, Liu, Cui, Zhou, Gong, Shen, Zhou, Chen, Gui, Zhang, and Huang]{gpt3survey}
Junjie Ye, Xuanting Chen, Nuo Xu, Can Zu, Zekai Shao, Shichun Liu, Yuhan Cui, Zeyang Zhou, Chao Gong, Yang Shen, Jie Zhou, Siming Chen, Tao Gui, Qi~Zhang, and Xuanjing Huang.
\newblock A comprehensive capability analysis of gpt-3 and gpt-3.5 series models, 2023.
\newblock URL \url{https://arxiv.org/abs/2303.10420}.

\bibitem[Rombach et~al.(2022)Rombach, Blattmann, Lorenz, Esser, and Ommer]{sd}
Robin Rombach, Andreas Blattmann, Dominik Lorenz, Patrick Esser, and Björn Ommer.
\newblock High-resolution image synthesis with latent diffusion models, 2022.
\newblock URL \url{https://arxiv.org/abs/2112.10752}.

\bibitem[Team et~al.(2025{\natexlab{b}})Team, Du, Gao, Xing, Jiang, Chen, Li, Xiao, Du, Liao, Tang, Wang, Zhang, Yuan, Lu, Tang, Sung, Wei, Lai, Guo, Zhu, Ding, Hu, Yang, Zhang, Yao, Zhao, Lu, Li, Yu, Gao, Zheng, Yuan, Chen, Guo, Su, Wang, Zhao, Zhang, Liu, Yan, Wu, Shi, Ye, Yu, Dong, Zhang, Ma, Pan, Gong, Liu, Ma, Wei, Cao, Huang, Jiang, Gao, Xiong, He, Huang, Xu, Wu, He, Wei, Jia, Wu, Xu, Zu, Zhou, Pan, Charles, Li, Hu, Liu, Chen, Wang, Liu, Qin, Liu, Yang, Bao, Du, Wu, Wang, Zhou, Wang, Li, Zhu, Zhang, Wang, Yang, Huang, Huang, Xu, Yang, and Lin]{kimi}
Kimi Team, Angang Du, Bofei Gao, Bowei Xing, Changjiu Jiang, Cheng Chen, Cheng Li, Chenjun Xiao, Chenzhuang Du, Chonghua Liao, Chuning Tang, Congcong Wang, Dehao Zhang, Enming Yuan, Enzhe Lu, Fengxiang Tang, Flood Sung, Guangda Wei, Guokun Lai, Haiqing Guo, Han Zhu, Hao Ding, Hao Hu, Hao Yang, Hao Zhang, Haotian Yao, Haotian Zhao, Haoyu Lu, Haoze Li, Haozhen Yu, Hongcheng Gao, Huabin Zheng, Huan Yuan, Jia Chen, Jianhang Guo, Jianlin Su, Jianzhou Wang, Jie Zhao, Jin Zhang, Jingyuan Liu, Junjie Yan, Junyan Wu, Lidong Shi, Ling Ye, Longhui Yu, Mengnan Dong, Neo Zhang, Ningchen Ma, Qiwei Pan, Qucheng Gong, Shaowei Liu, Shengling Ma, Shupeng Wei, Sihan Cao, Siying Huang, Tao Jiang, Weihao Gao, Weimin Xiong, Weiran He, Weixiao Huang, Weixin Xu, Wenhao Wu, Wenyang He, Xianghui Wei, Xianqing Jia, Xingzhe Wu, Xinran Xu, Xinxing Zu, Xinyu Zhou, Xuehai Pan, Y.~Charles, Yang Li, Yangyang Hu, Yangyang Liu, Yanru Chen, Yejie Wang, Yibo Liu, Yidao Qin, Yifeng Liu, Ying Yang, Yiping Bao, Yulun Du, Yuxin Wu, Yuzhi Wang, Zaida Zhou, Zhaoji Wang, Zhaowei Li, Zhen Zhu, Zheng Zhang, Zhexu Wang, Zhilin Yang, Zhiqi Huang, Zihao Huang, Ziyao Xu, Zonghan Yang, and Zongyu Lin.
\newblock Kimi k1.5: Scaling reinforcement learning with llms, 2025{\natexlab{b}}.
\newblock URL \url{https://arxiv.org/abs/2501.12599}.

\bibitem[Vaswani et~al.(2023)Vaswani, Shazeer, Parmar, Uszkoreit, Jones, Gomez, Kaiser, and Polosukhin]{attn}
Ashish Vaswani, Noam Shazeer, Niki Parmar, Jakob Uszkoreit, Llion Jones, Aidan~N. Gomez, Lukasz Kaiser, and Illia Polosukhin.
\newblock Attention is all you need, 2023.
\newblock URL \url{https://arxiv.org/abs/1706.03762}.

\bibitem[Shoeybi et~al.(2020)Shoeybi, Patwary, Puri, LeGresley, Casper, and Catanzaro]{mlm1}
Mohammad Shoeybi, Mostofa Patwary, Raul Puri, Patrick LeGresley, Jared Casper, and Bryan Catanzaro.
\newblock Megatron-lm: Training multi-billion parameter language models using model parallelism, 2020.
\newblock URL \url{https://arxiv.org/abs/1909.08053}.

\bibitem[Narayanan et~al.(2021)Narayanan, Shoeybi, Casper, LeGresley, Patwary, Korthikanti, Vainbrand, Kashinkunti, Bernauer, Catanzaro, Phanishayee, and Zaharia]{mlm2}
Deepak Narayanan, Mohammad Shoeybi, Jared Casper, Patrick LeGresley, Mostofa Patwary, Vijay~Anand Korthikanti, Dmitri Vainbrand, Prethvi Kashinkunti, Julie Bernauer, Bryan Catanzaro, Amar Phanishayee, and Matei Zaharia.
\newblock Efficient large-scale language model training on gpu clusters using megatron-lm, 2021.
\newblock URL \url{https://arxiv.org/abs/2104.04473}.

\bibitem[Korthikanti et~al.(2022)Korthikanti, Casper, Lym, McAfee, Andersch, Shoeybi, and Catanzaro]{mlm3}
Vijay Korthikanti, Jared Casper, Sangkug Lym, Lawrence McAfee, Michael Andersch, Mohammad Shoeybi, and Bryan Catanzaro.
\newblock Reducing activation recomputation in large transformer models, 2022.
\newblock URL \url{https://arxiv.org/abs/2205.05198}.

\bibitem[Liu et~al.(2025)Liu, Yan, Yao, Liu, Korthikanti, Wu, Fan, Deng, Bai, Chang, Aithal, Andersch, Shoeybi, Yao, Zhou, Wu, Li, and Yang]{mlm4}
Dennis Liu, Zijie Yan, Xin Yao, Tong Liu, Vijay Korthikanti, Evan Wu, Shiqing Fan, Gao Deng, Hongxiao Bai, Jianbin Chang, Ashwath Aithal, Michael Andersch, Mohammad Shoeybi, Jiajie Yao, Chandler Zhou, David Wu, Xipeng Li, and June Yang.
\newblock Moe parallel folding: Heterogeneous parallelism mappings for efficient large-scale moe model training with megatron core, 2025.
\newblock URL \url{https://arxiv.org/abs/2504.14960}.

\bibitem[Liu et~al.(2023)Liu, Zaharia, and Abbeel]{ringattn}
Hao Liu, Matei Zaharia, and Pieter Abbeel.
\newblock Ring attention with blockwise transformers for near-infinite context, 2023.
\newblock URL \url{https://arxiv.org/abs/2310.01889}.

\bibitem[Yang et~al.(2024)Yang, Zhang, Hui, Gao, Yu, Li, Liu, Tu, Zhou, Lin, Lu, Xue, Lin, Liu, Ren, and Zhang]{qwen25math}
An~Yang, Beichen Zhang, Binyuan Hui, Bofei Gao, Bowen Yu, Chengpeng Li, Dayiheng Liu, Jianhong Tu, Jingren Zhou, Junyang Lin, Keming Lu, Mingfeng Xue, Runji Lin, Tianyu Liu, Xingzhang Ren, and Zhenru Zhang.
\newblock Qwen2.5-math technical report: Toward mathematical expert model via self-improvement, 2024.
\newblock URL \url{https://arxiv.org/abs/2409.12122}.

\bibitem[DeepSeek-AI et~al.(2025{\natexlab{b}})DeepSeek-AI, Guo, Yang, Zhang, Song, Zhang, Xu, Zhu, Ma, Wang, Bi, Zhang, Yu, Wu, Wu, Gou, Shao, Li, Gao, Liu, Xue, Wang, Wu, Feng, Lu, Zhao, Deng, Zhang, Ruan, Dai, Chen, Ji, Li, Lin, Dai, Luo, Hao, Chen, Li, Zhang, Bao, Xu, Wang, Ding, Xin, Gao, Qu, Li, Guo, Li, Wang, Chen, Yuan, Qiu, Li, Cai, Ni, Liang, Chen, Dong, Hu, Gao, Guan, Huang, Yu, Wang, Zhang, Zhao, Wang, Zhang, Xu, Xia, Zhang, Zhang, Tang, Li, Wang, Li, Tian, Huang, Zhang, Wang, Chen, Du, Ge, Zhang, Pan, Wang, Chen, Jin, Chen, Lu, Zhou, Chen, Ye, Wang, Yu, Zhou, Pan, Li, Zhou, Wu, Ye, Yun, Pei, Sun, Wang, Zeng, Zhao, Liu, Liang, Gao, Yu, Zhang, Xiao, An, Liu, Wang, Chen, Nie, Cheng, Liu, Xie, Liu, Yang, Li, Su, Lin, Li, Jin, Shen, Chen, Sun, Wang, Song, Zhou, Wang, Shan, Li, Wang, Wei, Zhang, Xu, Li, Zhao, Sun, Wang, Yu, Zhang, Shi, Xiong, He, Piao, Wang, Tan, Ma, Liu, Guo, Ou, Wang, Gong, Zou, He, Xiong, Luo, You, Liu, Zhou, Zhu, Xu, Huang, Li, Zheng, Zhu, Ma, Tang, Zha, Yan, Ren, Ren, Sha, Fu, Xu, Xie, Zhang, Hao, Ma, Yan, Wu, Gu, Zhu, Liu, Li, Xie, Song, Pan, Huang, Xu, Zhang, and Zhang]{dseekr1}
DeepSeek-AI, Daya Guo, Dejian Yang, Haowei Zhang, Junxiao Song, Ruoyu Zhang, Runxin Xu, Qihao Zhu, Shirong Ma, Peiyi Wang, Xiao Bi, Xiaokang Zhang, Xingkai Yu, Yu~Wu, Z.~F. Wu, Zhibin Gou, Zhihong Shao, Zhuoshu Li, Ziyi Gao, Aixin Liu, Bing Xue, Bingxuan Wang, Bochao Wu, Bei Feng, Chengda Lu, Chenggang Zhao, Chengqi Deng, Chenyu Zhang, Chong Ruan, Damai Dai, Deli Chen, Dongjie Ji, Erhang Li, Fangyun Lin, Fucong Dai, Fuli Luo, Guangbo Hao, Guanting Chen, Guowei Li, H.~Zhang, Han Bao, Hanwei Xu, Haocheng Wang, Honghui Ding, Huajian Xin, Huazuo Gao, Hui Qu, Hui Li, Jianzhong Guo, Jiashi Li, Jiawei Wang, Jingchang Chen, Jingyang Yuan, Junjie Qiu, Junlong Li, J.~L. Cai, Jiaqi Ni, Jian Liang, Jin Chen, Kai Dong, Kai Hu, Kaige Gao, Kang Guan, Kexin Huang, Kuai Yu, Lean Wang, Lecong Zhang, Liang Zhao, Litong Wang, Liyue Zhang, Lei Xu, Leyi Xia, Mingchuan Zhang, Minghua Zhang, Minghui Tang, Meng Li, Miaojun Wang, Mingming Li, Ning Tian, Panpan Huang, Peng Zhang, Qiancheng Wang, Qinyu Chen, Qiushi Du, Ruiqi Ge, Ruisong Zhang, Ruizhe Pan, Runji Wang, R.~J. Chen, R.~L. Jin, Ruyi Chen, Shanghao Lu, Shangyan Zhou, Shanhuang Chen, Shengfeng Ye, Shiyu Wang, Shuiping Yu, Shunfeng Zhou, Shuting Pan, S.~S. Li, Shuang Zhou, Shaoqing Wu, Shengfeng Ye, Tao Yun, Tian Pei, Tianyu Sun, T.~Wang, Wangding Zeng, Wanjia Zhao, Wen Liu, Wenfeng Liang, Wenjun Gao, Wenqin Yu, Wentao Zhang, W.~L. Xiao, Wei An, Xiaodong Liu, Xiaohan Wang, Xiaokang Chen, Xiaotao Nie, Xin Cheng, Xin Liu, Xin Xie, Xingchao Liu, Xinyu Yang, Xinyuan Li, Xuecheng Su, Xuheng Lin, X.~Q. Li, Xiangyue Jin, Xiaojin Shen, Xiaosha Chen, Xiaowen Sun, Xiaoxiang Wang, Xinnan Song, Xinyi Zhou, Xianzu Wang, Xinxia Shan, Y.~K. Li, Y.~Q. Wang, Y.~X. Wei, Yang Zhang, Yanhong Xu, Yao Li, Yao Zhao, Yaofeng Sun, Yaohui Wang, Yi~Yu, Yichao Zhang, Yifan Shi, Yiliang Xiong, Ying He, Yishi Piao, Yisong Wang, Yixuan Tan, Yiyang Ma, Yiyuan Liu, Yongqiang Guo, Yuan Ou, Yuduan Wang, Yue Gong, Yuheng Zou, Yujia He, Yunfan Xiong, Yuxiang Luo, Yuxiang You, Yuxuan Liu, Yuyang Zhou, Y.~X. Zhu, Yanhong Xu, Yanping Huang, Yaohui Li, Yi~Zheng, Yuchen Zhu, Yunxian Ma, Ying Tang, Yukun Zha, Yuting Yan, Z.~Z. Ren, Zehui Ren, Zhangli Sha, Zhe Fu, Zhean Xu, Zhenda Xie, Zhengyan Zhang, Zhewen Hao, Zhicheng Ma, Zhigang Yan, Zhiyu Wu, Zihui Gu, Zijia Zhu, Zijun Liu, Zilin Li, Ziwei Xie, Ziyang Song, Zizheng Pan, Zhen Huang, Zhipeng Xu, Zhongyu Zhang, and Zhen Zhang.
\newblock Deepseek-r1: Incentivizing reasoning capability in llms via reinforcement learning, 2025{\natexlab{b}}.
\newblock URL \url{https://arxiv.org/abs/2501.12948}.

\bibitem[Cai et~al.(2025)Cai, Li, Li, Huang, Chen, Hu, and Wang]{dspo}
Miaomiao Cai, Simiao Li, Wei Li, Xudong Huang, Hanting Chen, Jie Hu, and Yunhe Wang.
\newblock Dspo: Direct semantic preference optimization for real-world image super-resolution, 2025.
\newblock URL \url{https://arxiv.org/abs/2504.15176}.

\bibitem[Schulman et~al.(2017)Schulman, Wolski, Dhariwal, Radford, and Klimov]{ppo}
John Schulman, Filip Wolski, Prafulla Dhariwal, Alec Radford, and Oleg Klimov.
\newblock Proximal policy optimization algorithms, 2017.
\newblock URL \url{https://arxiv.org/abs/1707.06347}.

\bibitem[Ramesh et~al.(2024)Ramesh, Hu, Chaimalas, Mehta, Sessa, Ammar, and Bogunovic]{grpo}
Shyam~Sundhar Ramesh, Yifan Hu, Iason Chaimalas, Viraj Mehta, Pier~Giuseppe Sessa, Haitham~Bou Ammar, and Ilija Bogunovic.
\newblock Group robust preference optimization in reward-free rlhf, 2024.
\newblock URL \url{https://arxiv.org/abs/2405.20304}.

\bibitem[Li et~al.(2024)Li, Xu, Zhang, Lin, Yu, Sun, and Luo]{remax}
Ziniu Li, Tian Xu, Yushun Zhang, Zhihang Lin, Yang Yu, Ruoyu Sun, and Zhi-Quan Luo.
\newblock Remax: A simple, effective, and efficient reinforcement learning method for aligning large language models, 2024.
\newblock URL \url{https://arxiv.org/abs/2310.10505}.

\bibitem[Zheng et~al.(2025)Zheng, Xing, Gu, Liang, Qu, Zhou, Li, Wen, Lin, Huang, et~al.]{ds_byte}
Tianyu Zheng, Tianshun Xing, Qingshui Gu, Taoran Liang, Xingwei Qu, Xin Zhou, Yizhi Li, Zhoufutu Wen, Chenghua Lin, Wenhao Huang, et~al.
\newblock First return, entropy-eliciting explore.
\newblock \emph{arXiv preprint arXiv:2507.07017}, 2025.

\bibitem[Shen et~al.(2024)Shen, Wang, Delalleau, Zeng, Dong, Egert, Sun, Zhang, Jain, Taghibakhshi, Ausin, Aithal, and Kuchaiev]{nemoaligner}
Gerald Shen, Zhilin Wang, Olivier Delalleau, Jiaqi Zeng, Yi~Dong, Daniel Egert, Shengyang Sun, Jimmy Zhang, Sahil Jain, Ali Taghibakhshi, Markel~Sanz Ausin, Ashwath Aithal, and Oleksii Kuchaiev.
\newblock Nemo-aligner: Scalable toolkit for efficient model alignment, 2024.

\bibitem[Hu et~al.(2024)Hu, Wu, Zhu, Xianyu, Wang, Zhang, and Cao]{openrlhf}
Jian Hu, Xibin Wu, Zilin Zhu, Xianyu, Weixun Wang, Dehao Zhang, and Yu~Cao.
\newblock Openrlhf: An easy-to-use, scalable and high-performance rlhf framework, 2024.
\newblock URL \url{https://arxiv.org/abs/2405.11143}.

\bibitem[Sheng et~al.(2025)Sheng, Zhang, Ye, Wu, Zhang, Zhang, Peng, Lin, and Wu]{verl}
Guangming Sheng, Chi Zhang, Zilingfeng Ye, Xibin Wu, Wang Zhang, Ru~Zhang, Yanghua Peng, Haibin Lin, and Chuan Wu.
\newblock Hybridflow: A flexible and efficient rlhf framework.
\newblock In \emph{Proceedings of the Twentieth European Conference on Computer Systems}, EuroSys ’25, page 1279–1297. ACM, March 2025.
\newblock \doi{10.1145/3689031.3696075}.
\newblock URL \url{http://dx.doi.org/10.1145/3689031.3696075}.

\bibitem[Li et~al.(2020{\natexlab{a}})Li, Zhao, Varma, Salpekar, Noordhuis, Li, Paszke, Smith, Vaughan, Damania, and Chintala]{torchddp}
Shen Li, Yanli Zhao, Rohan Varma, Omkar Salpekar, Pieter Noordhuis, Teng Li, Adam Paszke, Jeff Smith, Brian Vaughan, Pritam Damania, and Soumith Chintala.
\newblock Pytorch distributed: Experiences on accelerating data parallel training, 2020{\natexlab{a}}.
\newblock URL \url{https://arxiv.org/abs/2006.15704}.

\bibitem[Sergeev and Balso(2018)]{horovod}
Alexander Sergeev and Mike~Del Balso.
\newblock Horovod: fast and easy distributed deep learning in tensorflow, 2018.
\newblock URL \url{https://arxiv.org/abs/1802.05799}.

\bibitem[Huang et~al.(2019)Huang, Cheng, Bapna, Firat, Chen, Chen, Lee, Ngiam, Le, Wu, and Chen]{gpipe}
Yanping Huang, Youlong Cheng, Ankur Bapna, Orhan Firat, Mia~Xu Chen, Dehao Chen, HyoukJoong Lee, Jiquan Ngiam, Quoc~V. Le, Yonghui Wu, and Zhifeng Chen.
\newblock Gpipe: Efficient training of giant neural networks using pipeline parallelism, 2019.
\newblock URL \url{https://arxiv.org/abs/1811.06965}.

\bibitem[Harlap et~al.(2018)Harlap, Narayanan, Phanishayee, Seshadri, Devanur, Ganger, and Gibbons]{pipedream}
Aaron Harlap, Deepak Narayanan, Amar Phanishayee, Vivek Seshadri, Nikhil Devanur, Greg Ganger, and Phil Gibbons.
\newblock Pipedream: Fast and efficient pipeline parallel dnn training, 2018.
\newblock URL \url{https://arxiv.org/abs/1806.03377}.

\bibitem[Rajbhandari et~al.(2020{\natexlab{a}})Rajbhandari, Rasley, Ruwase, and He]{zero3}
Samyam Rajbhandari, Jeff Rasley, Olatunji Ruwase, and Yuxiong He.
\newblock Zero: Memory optimizations toward training trillion parameter models, 2020{\natexlab{a}}.
\newblock URL \url{https://arxiv.org/abs/1910.02054}.

\bibitem[Rajbhandari et~al.(2020{\natexlab{b}})Rajbhandari, Rasley, Ruwase, and He]{zero1}
Samyam Rajbhandari, Jeff Rasley, Olatunji Ruwase, and Yuxiong He.
\newblock Zero: Memory optimizations toward training trillion parameter models, 2020{\natexlab{b}}.
\newblock URL \url{https://arxiv.org/abs/1910.02054}.

\bibitem[Jacobs et~al.(2023)Jacobs, Tanaka, Zhang, Zhang, Song, Rajbhandari, and He]{ulysess}
Sam~Ade Jacobs, Masahiro Tanaka, Chengming Zhang, Minjia Zhang, Shuaiwen~Leon Song, Samyam Rajbhandari, and Yuxiong He.
\newblock Deepspeed ulysses: System optimizations for enabling training of extreme long sequence transformer models, 2023.
\newblock URL \url{https://arxiv.org/abs/2309.14509}.

\bibitem[Fang and Zhao(2024)]{fang2024uspunifiedsequenceparallelism}
Jiarui Fang and Shangchun Zhao.
\newblock Usp: A unified sequence parallelism approach for long context generative ai, 2024.
\newblock URL \url{https://arxiv.org/abs/2405.07719}.

\bibitem[Hwang et~al.(2023)Hwang, Cui, Xiong, Yang, Liu, Hu, Wang, Salas, Jose, Ram, Chau, Cheng, Yang, Yang, and Xiong]{tutel}
Changho Hwang, Wei Cui, Yifan Xiong, Ziyue Yang, Ze~Liu, Han Hu, Zilong Wang, Rafael Salas, Jithin Jose, Prabhat Ram, Joe Chau, Peng Cheng, Fan Yang, Mao Yang, and Yongqiang Xiong.
\newblock Tutel: Adaptive mixture-of-experts at scale, 2023.
\newblock URL \url{https://arxiv.org/abs/2206.03382}.

\bibitem[Bai et~al.(2022)Bai, Jones, Ndousse, Askell, Chen, DasSarma, Drain, Fort, Ganguli, Henighan, Joseph, Kadavath, Kernion, Conerly, El-Showk, Elhage, Hatfield-Dodds, Hernandez, Hume, Johnston, Kravec, Lovitt, Nanda, Olsson, Amodei, Brown, Clark, McCandlish, Olah, Mann, and Kaplan]{bai2022traininghelpfulharmlessassistant}
Yuntao Bai, Andy Jones, Kamal Ndousse, Amanda Askell, Anna Chen, Nova DasSarma, Dawn Drain, Stanislav Fort, Deep Ganguli, Tom Henighan, Nicholas Joseph, Saurav Kadavath, Jackson Kernion, Tom Conerly, Sheer El-Showk, Nelson Elhage, Zac Hatfield-Dodds, Danny Hernandez, Tristan Hume, Scott Johnston, Shauna Kravec, Liane Lovitt, Neel Nanda, Catherine Olsson, Dario Amodei, Tom Brown, Jack Clark, Sam McCandlish, Chris Olah, Ben Mann, and Jared Kaplan.
\newblock Training a helpful and harmless assistant with reinforcement learning from human feedback, 2022.
\newblock URL \url{https://arxiv.org/abs/2204.05862}.

\bibitem[Ouyang et~al.(2022)Ouyang, Wu, Jiang, Almeida, Wainwright, Mishkin, Zhang, Agarwal, Slama, Ray, Schulman, Hilton, Kelton, Miller, Simens, Askell, Welinder, Christiano, Leike, and Lowe]{ouyang2022traininglanguagemodelsfollow}
Long Ouyang, Jeff Wu, Xu~Jiang, Diogo Almeida, Carroll~L. Wainwright, Pamela Mishkin, Chong Zhang, Sandhini Agarwal, Katarina Slama, Alex Ray, John Schulman, Jacob Hilton, Fraser Kelton, Luke Miller, Maddie Simens, Amanda Askell, Peter Welinder, Paul Christiano, Jan Leike, and Ryan Lowe.
\newblock Training language models to follow instructions with human feedback, 2022.
\newblock URL \url{https://arxiv.org/abs/2203.02155}.

\bibitem[Bradley and Terry(1952)]{btrm}
Ralph~Allan Bradley and Milton~E Terry.
\newblock Rank analysis of incomplete block designs: I. the method of paired comparisons.
\newblock \emph{Biometrika}, 39\penalty0 (3/4):\penalty0 324--345, 1952.

\bibitem[Sun et~al.(2025)Sun, Shen, and Ton]{rethinkbt}
Hao Sun, Yunyi Shen, and Jean-Francois Ton.
\newblock Rethinking bradley-terry models in preference-based reward modeling: Foundations, theory, and alternatives, 2025.
\newblock URL \url{https://arxiv.org/abs/2411.04991}.

\bibitem[Mahan et~al.(2024)Mahan, Phung, Rafailov, Blagden, Lile, Castricato, Fränken, Finn, and Albalak]{genrm}
Dakota Mahan, Duy~Van Phung, Rafael Rafailov, Chase Blagden, Nathan Lile, Louis Castricato, Jan-Philipp Fränken, Chelsea Finn, and Alon Albalak.
\newblock Generative reward models, 2024.
\newblock URL \url{https://arxiv.org/abs/2410.12832}.

\bibitem[Wei et~al.(2023)Wei, Wang, Schuurmans, Bosma, Ichter, Xia, Chi, Le, and Zhou]{cot}
Jason Wei, Xuezhi Wang, Dale Schuurmans, Maarten Bosma, Brian Ichter, Fei Xia, Ed~Chi, Quoc Le, and Denny Zhou.
\newblock Chain-of-thought prompting elicits reasoning in large language models, 2023.
\newblock URL \url{https://arxiv.org/abs/2201.11903}.

\bibitem[Yu et~al.(2025)Yu, Zhang, Zhu, Yuan, Zuo, Yue, Dai, Fan, Liu, Liu, Liu, Lin, Lin, Ma, Sheng, Tong, Zhang, Zhang, Zhang, Zhu, Zhu, Chen, Chen, Wang, Yu, Song, Wei, Zhou, Liu, Ma, Zhang, Yan, Qiao, Wu, and Wang]{dapo}
Qiying Yu, Zheng Zhang, Ruofei Zhu, Yufeng Yuan, Xiaochen Zuo, Yu~Yue, Weinan Dai, Tiantian Fan, Gaohong Liu, Lingjun Liu, Xin Liu, Haibin Lin, Zhiqi Lin, Bole Ma, Guangming Sheng, Yuxuan Tong, Chi Zhang, Mofan Zhang, Wang Zhang, Hang Zhu, Jinhua Zhu, Jiaze Chen, Jiangjie Chen, Chengyi Wang, Hongli Yu, Yuxuan Song, Xiangpeng Wei, Hao Zhou, Jingjing Liu, Wei-Ying Ma, Ya-Qin Zhang, Lin Yan, Mu~Qiao, Yonghui Wu, and Mingxuan Wang.
\newblock Dapo: An open-source llm reinforcement learning system at scale, 2025.
\newblock URL \url{https://arxiv.org/abs/2503.14476}.

\bibitem[Yang et~al.(2025)Yang, Li, Yang, Zhang, Hui, Zheng, Yu, Gao, Huang, Lv, Zheng, Liu, Zhou, Huang, Hu, Ge, Wei, Lin, Tang, Yang, Tu, Zhang, Yang, Yang, Zhou, Zhou, Lin, Dang, Bao, Yang, Yu, Deng, Li, Xue, Li, Zhang, Wang, Zhu, Men, Gao, Liu, Luo, Li, Tang, Yin, Ren, Wang, Zhang, Ren, Fan, Su, Zhang, Zhang, Wan, Liu, Wang, Cui, Zhang, Zhou, and Qiu]{qwen3}
An~Yang, Anfeng Li, Baosong Yang, Beichen Zhang, Binyuan Hui, Bo~Zheng, Bowen Yu, Chang Gao, Chengen Huang, Chenxu Lv, Chujie Zheng, Dayiheng Liu, Fan Zhou, Fei Huang, Feng Hu, Hao Ge, Haoran Wei, Huan Lin, Jialong Tang, Jian Yang, Jianhong Tu, Jianwei Zhang, Jianxin Yang, Jiaxi Yang, Jing Zhou, Jingren Zhou, Junyang Lin, Kai Dang, Keqin Bao, Kexin Yang, Le~Yu, Lianghao Deng, Mei Li, Mingfeng Xue, Mingze Li, Pei Zhang, Peng Wang, Qin Zhu, Rui Men, Ruize Gao, Shixuan Liu, Shuang Luo, Tianhao Li, Tianyi Tang, Wenbiao Yin, Xingzhang Ren, Xinyu Wang, Xinyu Zhang, Xuancheng Ren, Yang Fan, Yang Su, Yichang Zhang, Yinger Zhang, Yu~Wan, Yuqiong Liu, Zekun Wang, Zeyu Cui, Zhenru Zhang, Zhipeng Zhou, and Zihan Qiu.
\newblock Qwen3 technical report, 2025.
\newblock URL \url{https://arxiv.org/abs/2505.09388}.

\bibitem[Kwon et~al.(2023)Kwon, Li, Zhuang, Sheng, Zheng, Yu, Gonzalez, Zhang, and Stoica]{vllm}
Woosuk Kwon, Zhuohan Li, Siyuan Zhuang, Ying Sheng, Lianmin Zheng, Cody~Hao Yu, Joseph~E. Gonzalez, Hao Zhang, and Ion Stoica.
\newblock Efficient memory management for large language model serving with pagedattention, 2023.
\newblock URL \url{https://arxiv.org/abs/2309.06180}.

\bibitem[Zheng et~al.(2024)Zheng, Yin, Xie, Sun, Huang, Yu, Cao, Kozyrakis, Stoica, Gonzalez, Barrett, and Sheng]{sglang}
Lianmin Zheng, Liangsheng Yin, Zhiqiang Xie, Chuyue Sun, Jeff Huang, Cody~Hao Yu, Shiyi Cao, Christos Kozyrakis, Ion Stoica, Joseph~E. Gonzalez, Clark Barrett, and Ying Sheng.
\newblock Sglang: Efficient execution of structured language model programs, 2024.
\newblock URL \url{https://arxiv.org/abs/2312.07104}.

\bibitem[Fu et~al.(2025)Fu, Gao, Shen, Zhu, Mei, He, Xu, Wei, Mei, Wang, Yang, Yuan, and Wu]{areal}
Wei Fu, Jiaxuan Gao, Xujie Shen, Chen Zhu, Zhiyu Mei, Chuyi He, Shusheng Xu, Guo Wei, Jun Mei, Jiashu Wang, Tongkai Yang, Binhang Yuan, and Yi~Wu.
\newblock Areal: A large-scale asynchronous reinforcement learning system for language reasoning, 2025.
\newblock URL \url{https://arxiv.org/abs/2505.24298}.

\bibitem[Xiao et~al.(2023)Xiao, Zhou, Mao, Wu, Zhao, Ju, Liang, Zhang, and Zhou]{flexrlhf}
Youshao Xiao, Zhenglei Zhou, Fagui Mao, Weichang Wu, Shangchun Zhao, Lin Ju, Lei Liang, Xiaolu Zhang, and Jun Zhou.
\newblock An adaptive placement and parallelism framework for accelerating rlhf training.
\newblock \emph{arXiv preprint arXiv:2312.11819}, 2023.

\bibitem[Xue et~al.(2025)Xue, Wu, Gao, Kong, Zhu, Chen, Liu, Liu, Guo, Huang, and Luo]{dancegrpo}
Zeyue Xue, Jie Wu, Yu~Gao, Fangyuan Kong, Lingting Zhu, Mengzhao Chen, Zhiheng Liu, Wei Liu, Qiushan Guo, Weilin Huang, and Ping Luo.
\newblock Dancegrpo: Unleashing grpo on visual generation, 2025.
\newblock URL \url{https://arxiv.org/abs/2505.07818}.

\bibitem[Deng and Raffel(2023)]{rad}
Haikang Deng and Colin Raffel.
\newblock Reward-augmented decoding: Efficient controlled text generation with a unidirectional reward model.
\newblock In \emph{Proceedings of the 2023 Conference on Empirical Methods in Natural Language Processing}, page 11781–11791. Association for Computational Linguistics, 2023.
\newblock \doi{10.18653/v1/2023.emnlp-main.721}.
\newblock URL \url{http://dx.doi.org/10.18653/v1/2023.emnlp-main.721}.

\bibitem[Li et~al.(2020{\natexlab{b}})Li, Song, Chen, Li, Liu, Tallent, and Barker]{evalgpuinter}
Ang Li, Shuaiwen~Leon Song, Jieyang Chen, Jiajia Li, Xu~Liu, Nathan~R. Tallent, and Kevin~J. Barker.
\newblock Evaluating modern gpu interconnect: Pcie, nvlink, nv-sli, nvswitch and gpudirect.
\newblock \emph{IEEE Transactions on Parallel and Distributed Systems}, 31\penalty0 (1):\penalty0 94–110, January 2020{\natexlab{b}}.
\newblock ISSN 2161-9883.
\newblock \doi{10.1109/tpds.2019.2928289}.
\newblock URL \url{http://dx.doi.org/10.1109/TPDS.2019.2928289}.

\bibitem[Zhang et~al.(2025)Zhang, Hosseini, Bansal, Kazemi, Kumar, and Agarwal]{genverifier}
Lunjun Zhang, Arian Hosseini, Hritik Bansal, Mehran Kazemi, Aviral Kumar, and Rishabh Agarwal.
\newblock Generative verifiers: Reward modeling as next-token prediction, 2025.
\newblock URL \url{https://arxiv.org/abs/2408.15240}.

\bibitem[Paszke et~al.(2019)Paszke, Gross, Massa, Lerer, Bradbury, Chanan, Killeen, Lin, Gimelshein, Antiga, Desmaison, Köpf, Yang, DeVito, Raison, Tejani, Chilamkurthy, Steiner, Fang, Bai, and Chintala]{torch}
Adam Paszke, Sam Gross, Francisco Massa, Adam Lerer, James Bradbury, Gregory Chanan, Trevor Killeen, Zeming Lin, Natalia Gimelshein, Luca Antiga, Alban Desmaison, Andreas Köpf, Edward Yang, Zach DeVito, Martin Raison, Alykhan Tejani, Sasank Chilamkurthy, Benoit Steiner, Lu~Fang, Junjie Bai, and Soumith Chintala.
\newblock Pytorch: An imperative style, high-performance deep learning library, 2019.
\newblock URL \url{https://arxiv.org/abs/1912.01703}.

\bibitem[Moritz et~al.(2018)Moritz, Nishihara, Wang, Tumanov, Liaw, Liang, Elibol, Yang, Paul, Jordan, and Stoica]{ray}
Philipp Moritz, Robert Nishihara, Stephanie Wang, Alexey Tumanov, Richard Liaw, Eric Liang, Melih Elibol, Zongheng Yang, William Paul, Michael~I. Jordan, and Ion Stoica.
\newblock Ray: A distributed framework for emerging ai applications, 2018.
\newblock URL \url{https://arxiv.org/abs/1712.05889}.

\bibitem[Graham et~al.(2005)Graham, Woodall, and Squyres]{openmpi}
Richard~L Graham, Timothy~S Woodall, and Jeffrey~M Squyres.
\newblock Open mpi: A flexible high performance mpi.
\newblock In \emph{International conference on parallel processing and applied mathematics}, pages 228--239. Springer, 2005.

\bibitem[Esposito and Haus(2025)]{slurm}
Aniello Esposito and Utz-Uwe Haus.
\newblock Slurm heterogeneous jobs for hybrid classical-quantum workflows, 2025.
\newblock URL \url{https://arxiv.org/abs/2506.03846}.

\bibitem[Wang et~al.(2025)Wang, Cai, Xie, Pan, Guan, Chu, Wang, Li, Huang, Cai, Hao, and Ding]{wlbllm}
Zheng Wang, Anna Cai, Xinfeng Xie, Zaifeng Pan, Yue Guan, Weiwei Chu, Jie Wang, Shikai Li, Jianyu Huang, Chris Cai, Yuchen Hao, and Yufei Ding.
\newblock Wlb-llm: Workload-balanced 4d parallelism for large language model training, 2025.
\newblock URL \url{https://arxiv.org/abs/2503.17924}.

\bibitem[Jiang et~al.(2024)Jiang, Lin, Zhong, Huang, Chen, Zhang, Peng, Li, Xie, Nong, Jia, He, Chen, Bai, Hou, Yan, Zhou, Sheng, Jiang, Xu, Wei, Zhang, Nie, Zou, Zhao, Xiang, Liu, Li, Jia, Ye, Jin, and Liu]{megascale}
Ziheng Jiang, Haibin Lin, Yinmin Zhong, Qi~Huang, Yangrui Chen, Zhi Zhang, Yanghua Peng, Xiang Li, Cong Xie, Shibiao Nong, Yulu Jia, Sun He, Hongmin Chen, Zhihao Bai, Qi~Hou, Shipeng Yan, Ding Zhou, Yiyao Sheng, Zhuo Jiang, Haohan Xu, Haoran Wei, Zhang Zhang, Pengfei Nie, Leqi Zou, Sida Zhao, Liang Xiang, Zherui Liu, Zhe Li, Xiaoying Jia, Jianxi Ye, Xin Jin, and Xin Liu.
\newblock Megascale: Scaling large language model training to more than 10,000 gpus, 2024.
\newblock URL \url{https://arxiv.org/abs/2402.15627}.

\bibitem[Team et~al.(2025{\natexlab{c}})Team, Kamath, Ferret, Pathak, Vieillard, Merhej, Perrin, Matejovicova, Ramé, Rivière, Rouillard, Mesnard, Cideron, bastien Grill, Ramos, Yvinec, Casbon, Pot, Penchev, Liu, Visin, Kenealy, Beyer, Zhai, Tsitsulin, Busa-Fekete, Feng, Sachdeva, Coleman, Gao, Mustafa, Barr, Parisotto, Tian, Eyal, Cherry, Peter, Sinopalnikov, Bhupatiraju, Agarwal, Kazemi, Malkin, Kumar, Vilar, Brusilovsky, Luo, Steiner, Friesen, Sharma, Sharma, Gilady, Goedeckemeyer, Saade, Feng, Kolesnikov, Bendebury, Abdagic, Vadi, György, Pinto, Das, Bapna, Miech, Yang, Paterson, Shenoy, Chakrabarti, Piot, Wu, Shahriari, Petrini, Chen, Lan, Choquette-Choo, Carey, Brick, Deutsch, Eisenbud, Cattle, Cheng, Paparas, Sreepathihalli, Reid, Tran, Zelle, Noland, Huizenga, Kharitonov, Liu, Amirkhanyan, Cameron, Hashemi, Klimczak-Plucińska, Singh, Mehta, Lehri, Hazimeh, Ballantyne, Szpektor, Nardini, Pouget-Abadie, Chan, Stanton, Wieting, Lai, Orbay, Fernandez, Newlan, yeong Ji, Singh, Black, Yu, Hui, Vodrahalli, Greff, Qiu, Valentine, Coelho, Ritter, Hoffman, Watson, Chaturvedi, Moynihan, Ma, Babar, Noy, Byrd, Roy, Momchev, Chauhan, Sachdeva, Bunyan, Botarda, Caron, Rubenstein, Culliton, Schmid, Sessa, Xu, Stanczyk, Tafti, Shivanna, Wu, Pan, Rokni, Willoughby, Vallu, Mullins, Jerome, Smoot, Girgin, Iqbal, Reddy, Sheth, Põder, Bhatnagar, Panyam, Eiger, Zhang, Liu, Yacovone, Liechty, Kalra, Evci, Misra, Roseberry, Feinberg, Kolesnikov, Han, Kwon, Chen, Chow, Zhu, Wei, Egyed, Cotruta, Giang, Kirk, Rao, Black, Babar, Lo, Moreira, Martins, Sanseviero, Gonzalez, Gleicher, Warkentin, Mirrokni, Senter, Collins, Barral, Ghahramani, Hadsell, Matias, Sculley, Petrov, Fiedel, Shazeer, Vinyals, Dean, Hassabis, Kavukcuoglu, Farabet, Buchatskaya, Alayrac, Anil, Dmitry, Lepikhin, Borgeaud, Bachem, Joulin, Andreev, Hardin, Dadashi, and Hussenot]{gemma3}
Gemma Team, Aishwarya Kamath, Johan Ferret, Shreya Pathak, Nino Vieillard, Ramona Merhej, Sarah Perrin, Tatiana Matejovicova, Alexandre Ramé, Morgane Rivière, Louis Rouillard, Thomas Mesnard, Geoffrey Cideron, Jean bastien Grill, Sabela Ramos, Edouard Yvinec, Michelle Casbon, Etienne Pot, Ivo Penchev, Gaël Liu, Francesco Visin, Kathleen Kenealy, Lucas Beyer, Xiaohai Zhai, Anton Tsitsulin, Robert Busa-Fekete, Alex Feng, Noveen Sachdeva, Benjamin Coleman, Yi~Gao, Basil Mustafa, Iain Barr, Emilio Parisotto, David Tian, Matan Eyal, Colin Cherry, Jan-Thorsten Peter, Danila Sinopalnikov, Surya Bhupatiraju, Rishabh Agarwal, Mehran Kazemi, Dan Malkin, Ravin Kumar, David Vilar, Idan Brusilovsky, Jiaming Luo, Andreas Steiner, Abe Friesen, Abhanshu Sharma, Abheesht Sharma, Adi~Mayrav Gilady, Adrian Goedeckemeyer, Alaa Saade, Alex Feng, Alexander Kolesnikov, Alexei Bendebury, Alvin Abdagic, Amit Vadi, András György, André~Susano Pinto, Anil Das, Ankur Bapna, Antoine Miech, Antoine Yang, Antonia Paterson, Ashish Shenoy, Ayan Chakrabarti, Bilal Piot, Bo~Wu, Bobak Shahriari, Bryce Petrini, Charlie Chen, Charline~Le Lan, Christopher~A. Choquette-Choo, CJ~Carey, Cormac Brick, Daniel Deutsch, Danielle Eisenbud, Dee Cattle, Derek Cheng, Dimitris Paparas, Divyashree~Shivakumar Sreepathihalli, Doug Reid, Dustin Tran, Dustin Zelle, Eric Noland, Erwin Huizenga, Eugene Kharitonov, Frederick Liu, Gagik Amirkhanyan, Glenn Cameron, Hadi Hashemi, Hanna Klimczak-Plucińska, Harman Singh, Harsh Mehta, Harshal~Tushar Lehri, Hussein Hazimeh, Ian Ballantyne, Idan Szpektor, Ivan Nardini, Jean Pouget-Abadie, Jetha Chan, Joe Stanton, John Wieting, Jonathan Lai, Jordi Orbay, Joseph Fernandez, Josh Newlan, Ju~yeong Ji, Jyotinder Singh, Kat Black, Kathy Yu, Kevin Hui, Kiran Vodrahalli, Klaus Greff, Linhai Qiu, Marcella Valentine, Marina Coelho, Marvin Ritter, Matt Hoffman, Matthew Watson, Mayank Chaturvedi, Michael Moynihan, Min Ma, Nabila Babar, Natasha Noy, Nathan Byrd, Nick Roy, Nikola Momchev, Nilay Chauhan, Noveen Sachdeva, Oskar Bunyan, Pankil Botarda, Paul Caron, Paul~Kishan Rubenstein, Phil Culliton, Philipp Schmid, Pier~Giuseppe Sessa, Pingmei Xu, Piotr Stanczyk, Pouya Tafti, Rakesh Shivanna, Renjie Wu, Renke Pan, Reza Rokni, Rob Willoughby, Rohith Vallu, Ryan Mullins, Sammy Jerome, Sara Smoot, Sertan Girgin, Shariq Iqbal, Shashir Reddy, Shruti Sheth, Siim Põder, Sijal Bhatnagar, Sindhu~Raghuram Panyam, Sivan Eiger, Susan Zhang, Tianqi Liu, Trevor Yacovone, Tyler Liechty, Uday Kalra, Utku Evci, Vedant Misra, Vincent Roseberry, Vlad Feinberg, Vlad Kolesnikov, Woohyun Han, Woosuk Kwon, Xi~Chen, Yinlam Chow, Yuvein Zhu, Zichuan Wei, Zoltan Egyed, Victor Cotruta, Minh Giang, Phoebe Kirk, Anand Rao, Kat Black, Nabila Babar, Jessica Lo, Erica Moreira, Luiz~Gustavo Martins, Omar Sanseviero, Lucas Gonzalez, Zach Gleicher, Tris Warkentin, Vahab Mirrokni, Evan Senter, Eli Collins, Joelle Barral, Zoubin Ghahramani, Raia Hadsell, Yossi Matias, D.~Sculley, Slav Petrov, Noah Fiedel, Noam Shazeer, Oriol Vinyals, Jeff Dean, Demis Hassabis, Koray Kavukcuoglu, Clement Farabet, Elena Buchatskaya, Jean-Baptiste Alayrac, Rohan Anil, Dmitry, Lepikhin, Sebastian Borgeaud, Olivier Bachem, Armand Joulin, Alek Andreev, Cassidy Hardin, Robert Dadashi, and Léonard Hussenot.
\newblock Gemma 3 technical report, 2025{\natexlab{c}}.
\newblock URL \url{https://arxiv.org/abs/2503.19786}.

\bibitem[Bai et~al.(2025)Bai, Chen, Liu, Wang, Ge, Song, Dang, Wang, Wang, Tang, Zhong, Zhu, Yang, Li, Wan, Wang, Ding, Fu, Xu, Ye, Zhang, Xie, Cheng, Zhang, Yang, Xu, and Lin]{qwen25vl}
Shuai Bai, Keqin Chen, Xuejing Liu, Jialin Wang, Wenbin Ge, Sibo Song, Kai Dang, Peng Wang, Shijie Wang, Jun Tang, Humen Zhong, Yuanzhi Zhu, Mingkun Yang, Zhaohai Li, Jianqiang Wan, Pengfei Wang, Wei Ding, Zheren Fu, Yiheng Xu, Jiabo Ye, Xi~Zhang, Tianbao Xie, Zesen Cheng, Hang Zhang, Zhibo Yang, Haiyang Xu, and Junyang Lin.
\newblock Qwen2.5-vl technical report, 2025.
\newblock URL \url{https://arxiv.org/abs/2502.13923}.

\bibitem[Zheng et~al.(2017)Zheng, Lin, Xu, Wei, Zeng, Yang, and Zhang]{wxpaxos}
Jianjun Zheng, Qian Lin, Jiatao Xu, Cheng Wei, Chuwei Zeng, Pingan Yang, and Yunfan Zhang.
\newblock Paxosstore: high-availability storage made practical in wechat.
\newblock \emph{Proceedings of the VLDB Endowment}, 10\penalty0 (12):\penalty0 1730--1741, 2017.

\bibitem[Cobbe et~al.(2021)Cobbe, Kosaraju, Bavarian, Chen, Jun, Kaiser, Plappert, Tworek, Hilton, Nakano, Hesse, and Schulman]{gsm8k}
Karl Cobbe, Vineet Kosaraju, Mohammad Bavarian, Mark Chen, Heewoo Jun, Lukasz Kaiser, Matthias Plappert, Jerry Tworek, Jacob Hilton, Reiichiro Nakano, Christopher Hesse, and John Schulman.
\newblock Training verifiers to solve math word problems, 2021.
\newblock URL \url{https://arxiv.org/abs/2110.14168}.

\end{thebibliography}

\end{document}